\RequirePackage{fix-cm}
\documentclass[smallcondensed]{svjour3}     % %onecolumn(ditto)
\smartqed  % flush right qed marks, e.g. at end of proof
\usepackage[T1]{fontenc}
\usepackage{lmodern}      %Added on 26.08.2022 Dina
\usepackage{amsmath}
\usepackage{graphicx}
\usepackage{float}
\usepackage{xcolor}
\usepackage{dsfont}
\usepackage{mathrsfs}

\usepackage{mathrsfs} %pour les lettres majuscules rondes ==> fonction \mathscr 

\usepackage[gen]{eurosym}
\usepackage{verbatim}
\usepackage{amssymb}
\usepackage{latexsym}		% Symboles mathmatiques
\usepackage{tabularx}
\usepackage{setspace}
\usepackage{listings}

\usepackage[utf8]{inputenc} % allow utf-8 input
\usepackage{hyperref}       % hyperlinks
\usepackage{url}            % simple URL typesetting
\usepackage{booktabs}       % professional-quality tables
\usepackage{amsfonts}       % blackboard math symbols
\usepackage{nicefrac}       % compact symbols for 1/2, etc.
\usepackage{microtype}      % microtypography

\usepackage{float}
\usepackage{multirow,lscape,longtable}
\usepackage[numbers]{natbib}

 \journalname{European Actuarial Journal}

\begin{document}

\title{Model Transparency and Interpretability : Survey and Application to the Insurance Industry%\thanks{Grants or other notes
%about the article that should go on the front page should be
%placed here. General acknowledgments should be placed at the end of the article.}
}
%\subtitle{Do you have a subtitle?\\ If so, write it here}

\titlerunning{Model Transparency and Interpretability}        % if too long for running head

\author{Dimitri Delcaillau      \and
        Antoine Ly \and
        Alize Papp \and
        Franck Vermet}

%\authorrunning{Short form of author list} % if too long for running head

\institute{Dimitri Delcaillau \at
              P\&C, Predictive Analytics\\
  Milliman\\
  Paris, France \\
\email{dimitri.delcaillau@milliman.com}           
\and
Antoine Ly \at
 Data Analytics Solutions\\
  SCOR \\
  Paris, France \\
  \email{aly@scor.com} 
  \and
   Aliz\'e Papp \at
   Data Analytics Solutions\\
  SCOR \\
  Charlotte, USA \\
  \email{alize.papp@outlook.com}
  \and
  Franck Vermet \at
  Laboratoire de Math\'ematiques de Bretagne Atlantique, EURIA \\                          
  Univ Brest\\
  Brest, France \\
  \email{franck.vermet@univ-brest.fr} \\
  ORCID Number : 0000-0003-3816-5401 \\
}
\date{Received: date / Accepted: date}
% The correct dates will be entered by the editor

\maketitle

\begin{abstract}
%Insert your abstract here. Include keywords, PACS and mathematical subject classification numbers as needed.
%\input{Introduction.tex}
The use of models, even if efficient, must be accompanied by an understanding at all levels of the process that transforms data (upstream and downstream). Thus, needs increase to define the relationships between individual data and the choice that an algorithm could make based on its analysis (e.g. the recommendation of one product or one promotional offer, or an insurance rate representative of the risk). Model users must ensure that models do not discriminate and that it is also possible to explain their results. 
This paper introduces the importance of model interpretation and tackles the notion of model transparency. Within an insurance context, it specifically illustrates how some tools can be used to enforce the control of actuarial models that can nowadays leverage on machine learning. On a simple example of loss frequency estimation in car insurance, we show the interest of some interpretability methods to adapt explanation to the target audience.

% \PACS{PACS code1 \and PACS code2 \and more}
% \subclass{MSC code1 \and MSC code2 \and more}
\end{abstract}
\newpage

\section{Purpose of this paper} 
Since the General Data Protection Regulation (GDPR) was released in May 2018, regulations have been actively working on controlling the right usage of data and models. The insurance industry is regulated for many years and historically allocates high importance on model control, critical for insurance products. The use of algorithms is made with a strong will to have an efficient and accurate estimation of risks but also needs to understand the outcome. Model users  and developer must ensure algorithms explainability and non-discriminatory decision. Because there are more and more predictive algorithms -- which is made possible by the evolution of computing capacities -- scientists must be vigilant about the use of their models and to consider new tools to better understand the decisions they can suggest taking. Recently, the community has been particularly active on model transparency with a marked intensification of publications over the past three years. The increasingly frequent use of more complex algorithms (deep learning, Xgboost, etc.) which have attractive performance is undoubtedly one of the roots of this interest.\\

 This paper focuses on the importance of model interpretation and model transparency. It specifically illustrates the usage of certain tools within an actuarial context to enforce the control of predictive models that can nowadays leverage on machine learning. We present, how some tools on transparency can be used to face the challenges of controlling the "black box". We especially try to share a definition of this notion and how it can be assessed on models used in insurance and answer to some questions that specific audiences might have. We acknowledge that many methodologies exist to help users building interpretable models and research is very active on this topic. We focus on the most popular methodologies and illustrate their application on a insurance pricing use case. The paper is structured as followed:
\begin{itemize}
    \item \textbf{First section}: explaining challenges of model interpretability and transparency. We share a definition that guides our choice of tools we illustrate in an insurance context.
    \item \textbf{Second section}: presenting an overview of a selection of transparency tools that could help actuaries or people exposed to modeling in insurance. We highlight pros and cons of such tools and how they could help.
    \item \textbf{Last section}: a concrete application in insurance by taking the example of an actuarial pricing. Transparency tools are used and presented in regards to a specific audience that models are susceptible to be explained to. \\
\end{itemize}

Finally, model interpretability is often linked to "Ethics or Fairness" which is not tackled in this paper. The choice of examples is made to illustrate some key concepts and how they could be applied to insurance topics. We invite the reader to look at other methodologies (some of which are recommended in the references and appendix) as this article is simply sharing a position based on research we made and industry experience.

\section{Model interpretability: a major challenge} 
\label{sec:0_interpretabilite}
\subsection{Defining interpretability}

%\begin{figure}[H]
%\begin{center}
%\includegraphics[scale=0.39]{pics/nb_articles_iml.png}
%\end{center}
%\caption{Number of published articles on the interpretability of machine learning models (in the last 15 years)}
%    \label{fig:nb_papers_interpretability}
%\end{figure}

The interpretability of models is a new issue in the use of models and more particularly those of machine learning, as publications of the last  years can attest. \cite{Peeking_Black_Box_XAI} have indeed shown the growing interest of the scientific community and regulators in model interpretation. Surprisingly enough, although the concept of interpretability is increasingly widespread, there is no general consensus on both the definition and the measurement of the interpretability of a model \cite{molnar2019}. There are indeed many methods (graphs, mathematics, etc.) that can be associated with the interpretation of algorithms, which sometimes leads to confusion around the notion. Moreover, the term "interpretation" can refer to different degrees of understanding depending on the target population: are we talking of understanding of the model, controlling its results, its transparency towards novice users? Or are we referring to the means put in place to analyze the results of an algorithm, however complex it may be? \\ 

Murdoch et al. \cite{define_interpretability_ML} try to give a precise definition of interpretability within the framework of a machine learning model. In particular, they provide a framework (called \textit{PDR}) built on three desired properties for the evaluation and construction of an interpretation: predictive accuracy (\textit{P}), descriptive accuracy (\textit{D}) and relevance (\textit{R}). This allows, among other things, to classify the different existing methods and to use a common vocabulary between the different actors in the field of machine learning. \cite{miller} proposes more specifically to define interpretability as the degree to which a human can understand the cause of a decision. An alternative definition is also proposed by \cite{NIPS2016_6300}, which \cite{molnar2019} also endorses: interpretability is defined as \guillemotleft \textbf{the degree to which a human can consistently predict the model's result}\guillemotright. If having a unique definition is difficult, there is some consensus that leads to assume that information is said to be relevant if it provides insights for a particular audience and a problem in a chosen domain. \\
\cite{miller} and \cite{molnar2019} will strongly guide the definition we adopt to model interpretability and the way it can be applied to assess insurance model relevance.

\subsection{The criteria of an interpretable model}

Based on the definition in \cite{miller} and \cite{molnar2019}, the notion of interpretability can be translated according to several criteria:

\begin{itemize}
    \item \textbf{Trust}: this criterion comes up regularly when the interpretability of models is discussed. For example, \cite{lime} explains that if a model provides results a human uses to make decisions, it is clear that he or she must be able to rely on the model with complete peace of mind. 
    \item \textbf{Causality}: although one of the objectives of machine learning algorithms is to highlight correlations, the algorithm must provide a better understanding of real-world phenomena and the interactions between different observed factors.
    \item \textbf{\guillemotleft Transferability\guillemotright}: defined as the ability of a model to adapt to slightly different situations, it is one of the desired properties in the search for interpretability. In addition, it transcribes the capacity for generalization.
    \item \textbf{\guillemotleft Informativity\guillemotright}: the use of an algorithm must be able to go beyond simple mathematical optimization. A model must be able to provide precise information on its decision making.
    \item \textbf{Fair and ethical decision making}: this criterion is in line with the guidelines of the GDPR. The user of an algorithm must be able to ensure that there is no bias in the decision making and that it is ethical (e.g., it does not discriminate). While this paper does not tackle precisely these criteria, it presents at least model interpretability as a critical topic to assess model fairness.
\end{itemize}

%\textcolor{red}{PDR to be removed as we don't apply it}
Even if it is difficult to objectively take into account all of these axes, two main approaches can help to guarantee interpretability: those based on the model itself, and those that apply \textit{a posteriori} via \textit{post-hoc} analyses. %In order to choose the proper approach, the PDR framework suggests three criteria: \guillemotleft predictive accuracy\guillemotright, \guillemotleft descriptive accuracy\guillemotright and \guillemotleft relevance\guillemotright.

\subsection{The two main types of interpretability}

\subsubsection{Model-Based Interpretability (MBI)}

The first level of interpretability is model-based interpretability (MBI). It comes into play during the development of the model and is linked to both the choice of families of algorithms used to understand a phenomenon, and their calibration. From this point of view, an interpretable model can be defined by its :

\begin{itemize}
    \item \textbf{Sparsity}: Sparsity is closely related to Occam's razor principle, which states that multiples should not be used unnecessarily. In the case of machine learning, that means limiting the number of non-zero parameters. In statistics as in machine learning, there are different regularization methods \footnote{Xgboost introduces a regularization method just like the drop-out in deep-learning. However, even under constraint, these models are often not very sparse.}. 
    \item \textbf{Simulatability}: \cite{define_interpretability_ML} defines a model as simulatable if a human is able to internally simulate the entire decision process of the algorithm. Thus, simulatability refers to an almost full transparency of the model: a human should be able, from the inputs and parameters of the model, to perform all the necessary calculations, in reasonable time, and find the model's prediction. In this sense, decision trees are generally cited as simulatable algorithms, given their visual simplicity for decision making. Similarly, decision rules fall into this category.
    \item \textbf{Modularity}: a model is modular if a significant portion of the prediction process can be interpreted independently. Thus, a modular model will not be as simple to understand as a sparse or simulatable model but it can increase descriptive accuracy by providing relationships learned by the algorithm. A classical example of a model considered as modular is the family of GAMs (Generalized Additive Models) \cite{tibshirani1990generalized}, of which GLMs are a subfamily. In this type of models, the relationship between variables is necessarily additive and the calibrated coefficients allow for a relatively easy interpretation of the model. On the other hand, deep neural networks are considered as not very modular, given the little information their coefficients provide for each layer. For instance, in a study by Caruana et al. (2015), it is shown that the probability of death due to pneumonia is lower for patients with asthma. This is because these patients receive more aggressive treatment; but if a doctor followed the algorithm's recommendations and gave a less aggressive treatment for people with asthma, the model would be dangerously wrong. This example shows the value of modularity to produce relevant interpretations, so that biases in the learning dataset can be detected afterwards.
\end{itemize}

 %According to its nature, a model has properties that can allow for a more simple understanding of its inner workings. On the contrary, the second level of interpretation is less sensitive to the algorithms used in the modeling process.

\subsubsection{\textit{Post-hoc} interpretability}

Post-hoc interpretability, unlike model-based interpretability, corresponds to an analysis after the model has been fitted. This a posteriori interpretation occurs in order to provide information on the possible relationships the algorithm might have captured. It is on this type of interpretation that research has been particularly active in recent years. 

Post-hoc methods have proven to be particularly useful for analyzing complex models with high predictive accuracy: the goal is to increase descriptive accuracy while maintaining predictive accuracy constant. There are two levels: understanding the model with regard to the data it was trained on, and analyzing the model's predictions. Therefore it adds a new layer to the model. In recent years, these methods have vastly expanded, making it possible to overcome the limits of pre-existing analysis tools, particularly those based on trees \cite{Breiman2001} or neural networks \cite{olden2004accurate}. \\

Interpretation at the data level allows to focus on the general relationships the model learned, i.e. the relevant rules for a particular class of responses or sub-population. Both of these levels of post-hoc interpretation have global and local interpretation tools. The next section introduces different methods of post-hoc interpretation.

\section{Overview on some interpretation methods \textit{post-hoc}} 
\label{sec:1_methodesInterpretation}

\begin{figure}[H]
    \centering
    \includegraphics[scale=0.7]{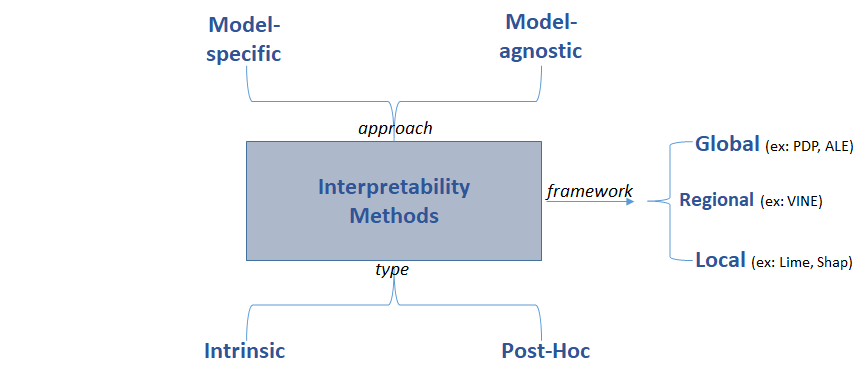}
    \caption{The different categories of interpretation methods}
    \label{scope_interpretability}
\end{figure}

In this section, we focus on some post-hoc and model agnostic interpretation methods.
Of course, there are other methods, which are specific to each algorithm ; these methods may reinforce the interpretation of the model (such as the importance of features in trees \cite{Breiman2001}) but we choose not to describe them here, to not limit the generality of the study. The Figure \ref{scope_interpretability} divides interpretation methods according to their type, their approach and their framework. 
In what follows, we focus on the frameworks: the literature comprises two main families, local and global methods. While local methods aim at understanding the black-box model prediction for a particular observation, the global approach tries to understand the model as a whole. Halfway between these two families is the regional framework, which tries to explain the behavior of the model for a group of similar observations, for example from clusters. In the following subsections, we detail the interpretation tools that seem the most relevant in the context of the insurance industry and the complex models that are often used there, that is tree models (Random Forest or Gradient Boosting). 
Note, however, that these methods are also applicable to any algorithm (which is why they are called model-"agnostic").

Note also that given the amount of existing methods to interpret a machine learning model, we had to select some of them in order to suggest a general method that can be applied in practice.
The following graph introduces a path, from the most general to the most precise, to study and interpret a model, and some examples of methods for each category.

\begin{figure}[H]
    \centering
    \includegraphics[scale=0.7]{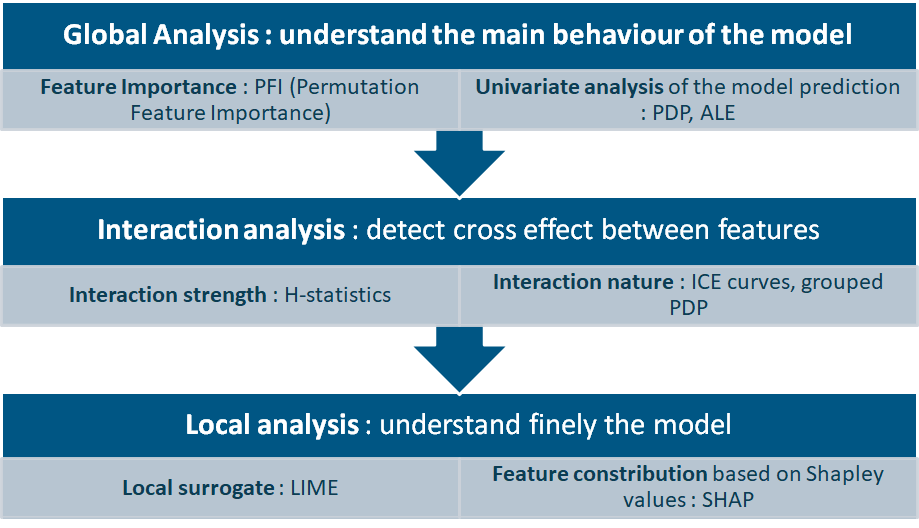}
    \caption{Suggested methodology to interpret a machine learning model}
    \label{frame_interpretability}
\end{figure}

The next subsections describe the selected methods, from the theory to the advantages and limits of each one.

\subsection{Global methods}

When we fit a complex model, the first information we want to identity is its main behaviour and the global role of each feature in the prediction.
As the name implies, the techniques to get these pieces of information are placed in the category of global methods.
We decided to focus on two aspects of global methods : the first one is to determine which features contribute less and more in the model (feature importance methods); the second one is to understand the average marginal effect of the features on the model prediction.

Note that we can also mention another approach : methods providing global surrogate models such as TREPAN \cite{trepan} or Born Again Trees \cite{Breiman1996} for instance give a global explanation of the model. Those methods allow measuring the fidelity between the surrogate model (that serves for explanation) and the black box. The idea is to build an interpretable decision tree that mimics the complex model by generating artificial samples with the complex model calibrated on the observed data. In fact, we can read in \cite{first_inter} that first methods of interpretability tried to extract sets of rules or trees from neural networks~: global surrogates were probably the first approach for interpretability.

\subsubsection{Features Importance}
Some of them are specific to a model or a class of models: $t$-statistics is an example in the case of linear models but there are also specific measures for tree-based models. Here, we are interested in a new definition of the importance of variables, which has the particularity of being independent of the model under consideration, noted PFI (\emph{Permutation Feature Importance}) \cite{variable_importance}.

The idea is to consider that if a feature is very important, the alteration of the quality of its value will greatly disturb the quality of the model predictions. To do this, we artificially alter the information for this feature by swapping all its values. If the prediction of a model is greatly modified when the values of a variable are swapped, it means that the model is sensitive to variations in this variable and therefore plays a preponderant role in the model. Conversely, a variable for which a change in its values will have little impact on the model prediction will not be considered important. In summary, a variable is all the more important as the prediction error of the model increases after having swapped the values of this variable.

Let us describe the calculation of this statistic. Let $A_n=\{Z^{(i)}=(X^{(i)},Y^{(i)}) \in \mathds{R}^p$x$\mathds{R}, i=1,\dots, n\}$ be a training set, which consists of $n$ independent random vectors following the same law and a model $\hat{f}$, trained on observations  $(x^{(i)},y^{(i)})$ of the set $A_n$, where $x^{(i)}=(x_j^{(i)})_{1 \leq j \leq p}$ for $\ 1  \leq i \leq n$.  Denote by $L$ the error function used, for example: $L(y,\hat{f}(x)) =\frac{1}{n} \sum\limits_{i=1}^n{(y^{(i)}-\hat{f}(x^{(i)}))^2}$.
The procedure for calculating the importance of the $p$ variables in the model is as follows:

\begin{itemize}
    \item Calculation of the original error of the model: $err_1=L(y,\hat{f}(x))$
    \item for $j=1$ to $p$ :
    \begin{itemize}
        \item We randomly choose a  permutation $\sigma$ of $\{1,...,n\}$ in $\{1,...,n\}$.
       We choose a new matrix $(\tilde{x}_j^{(i)})_{\underset{1 \leq i \leq n}{1 \leq j \leq p}}$ of input variables, by the formula :
        $$\forall i \in \{1,...,n\}, \forall k \in \{1,...,p\}, \tilde{x}_k^{(i)}= \left\{\begin{matrix}
x_k^{(i)} \ si \ k \neq j \\ 
x_j^{(\sigma(i))} \ si \ k=j
\end{matrix}\right.$$\\
That is to say, we permute (with $\sigma$) the $n$ observations of the variable $x_j$ and leave the other observations unchanged.
\item We estimate the error made by the model on this new input matrix, namely: $err_j=L(y,\hat{f}(\tilde{x}))$.
\item The importance of the variable $x_j$ is calculated by the formula: $FI^{(j)}=err_j/err_1$
    \end{itemize}
\item Output of the algorithm: $FI^{(1)},...,FI^{(p)}$, sorted in descending order.
\end{itemize}

Depending on the type of interpretation one wishes to obtain, it may be more relevant to calculate the importance of variables on the learning basis $B_a$ or the test basis $B_t$: the calculation on $B_a$ makes it possible to know how much the model relies on each variable to make a prediction; the calculation on $B_t$ makes it possible to know how much a variable contributes to the performance of the model on untrained data.

\paragraph{Advantages}
The advantages of this method of measuring the importance of variables are manifold, including the following:
\begin{itemize}
    \item An intuitive measure: the greater the error when the information is deteriorated, the greater the variable.
    \item A synthetic global overview, as one could have with the coefficients of a linear regression model.
    \item A criterion that is comparable between different models.
    \item A consideration of both the effects of the variable and its interactions with other variables\footnote{This can also be seen as a drawback.}.
    \item An efficient calculation that does not require the model to be re-trained, which saves time compared to other methods.
\end{itemize}

\paragraph{Disadvantages}
Importance by permutation has, however, some disadvantages:
\begin{itemize}
    \item The result provided by the algorithm can vary greatly due to the randomness introduced by the permutations.
    \item The addition of a correlated variable to another decreases the importance of the variable under consideration.
    \item The permutations can provide unrealistic instances. Indeed, when we permute a variable within an instance, we do not pay attention to the fact that the new instance is really observable. This is the same problem as the one observed with the PDP. Let us consider for example the case where we have the explanatory variables of the weight and height of a man. If we perform a permutation as in the above algorithm, we can end up with an individual of height 2 meters and weight 30kg, which is not possible in reality.
\end{itemize}

We decided to not go into detail for this method, but the principle behind is to consider that a feature  is all the more important as the prediction error of the model increases after permuting the values of the feature under consideration.

We are able to understand which features contribute the most in a fitted model, we will now see techniques to determine how the model responds according to a specific feature.

\subsubsection{Partial Dependence Plots (PDP)}
\label{subsec:PDP}

\paragraph{Presentation of the method}
PDP analysis (Partial Dependence Plot), was introduced by \cite{pdp}. This  global interpretation method is a graphical method, which  aims to show the marginal effect of one or more explanatory variables on the prediction made by a model.\\

Let $A_n=\{Z^{(i)}=(X^{(i)},Y^{(i)}) \in \mathds{R}^p$x$\mathds{R}, i=1,\dots, n\}$ be a training set, which consists of $n$ independent random vectors following the same law and a model $\hat{f}$, trained on observations  $(x^{(i)},y^{(i)})$ of the set $A_n$, where $x^{(i)}=(x_j^{(i)})_{1 \leq j \leq p}$ for $\ 1  \leq i \leq n$ . Let $X_S$ be the set of variables we want to know the effect on the prediction, and $X_C$ the other explanatory variables. For instance, $X_S=(X_1,X_2)$ and $X_C=(X_3,...,X_p)$. Hence, $X=(X_S,X_C)$, denote the set of explanatory variables in the model.\\

The partial dependence function is then defined as:
\begin{equation}\hat{f}_{x_S}(x_S)=\mathds{E}_{X_c}[\hat{f}(x_S,X_C)] = \int\hat{f}(x_s,x_c)d\mathds{P}_{X_c}(x_c).\end{equation}

Note this formula is not equivalent to the conditional expectation of $X_s$. To estimate the PDP function, we can just use the $n$ observations and the Monte-Carlo method to estimate the expected value, i.e.:

\begin{equation}\hat{f}_{x_S}(x_S) \simeq \frac{1}{n} \sum\limits_{i=1}^{n}{\hat{f}(x_S,x_C^{(i)})}. \end{equation}

%\begin{figure}[!h]
%    \centering
%    \includegraphics[width=120mm]{pics/PD_calcul.png}
%    \caption{Computation of the PDP plot for a simple example}
%    \label{fig:PD_calcul}
%\end{figure}

The algorithm to build the PDP plot is detailed out in Appendix \ref{algo_pdp}. It is based on the strong assumption of non-correlation between the variables in the $C$ set and those in $S$. In practice, this case is rarely verified, which leads to the consideration of associations of modalities that are not possible in reality (for example, observing a 2-meter tall individual with a weight of less than 10kg, cf. Section \ref{subsec:ALE}).\\

\begin{figure}[!h]
\centering
\includegraphics[width=80mm]{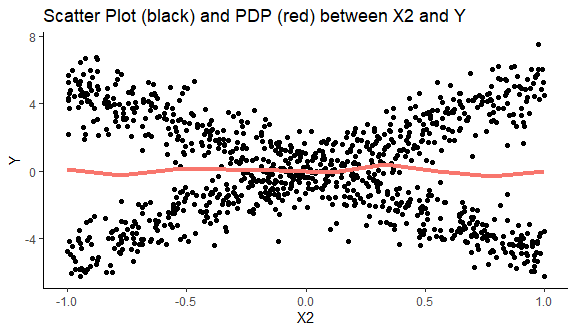} 
\caption{Scatter plot of $X_2$ and $Y$ (black) and PDP of $X_2$ (red)}
    \label{fig:scatterPDP}
\end{figure}

\paragraph{Example}
To illustrate that method, let us consider the following example. We have variables that are linked in the following way:

$$Y=0.2X_1-5X_2+10 X_2 \mathds{1}_{\{ X_3 \geq 0\}}+ \varepsilon,$$ where $X_1,X_2,X_3 \overset{iid}{\sim}{U(-1,1)}, \varepsilon \sim N(0,1) $ with $\varepsilon$ independent of $X_1,X_2,X_3$.
Assume we have a sample of size $n=1000$.
The scatter plot of $X_2$ and $Y$ for this sample is shown in black in Figure \ref{fig:scatterPDP}. We train a random forest model to predict $Y$. The PDP plot for $X_2$ is shown in red in the same figure.

\

The PDP plot of $X_2$ is not predictive on $Y$ on average, when the scatter plot leads to the opposite conclusion. This example illustrates the problem of not taking into account the correlations between the variables. To take into account the dependency, there is an alternative to PDP called ALE, which is detailed in the next subsection.\\

\paragraph{\textbf{Conclusion: contributions and limitations of PDP.}}

PDP plots are often used for their simplicity of interpretation and ease of implementation. In addition, these plots can be used as a tool in estimating the importance of variables and their interactions within a model. However, this graph alone is not sufficient to explain all the complexity of an algorithm. The calculation method is indeed based on the strong and limiting assumption of independence between variables. Moreover, PDP masks heterogeneous effects as shown in Figure \ref{fig:scatterPDP}. This is the reason why this method is often associated with other graphs such as the ICE detailed in Appendix \ref{ICE_methodo}.

\subsubsection{Accumulated Local Effects Plots (ALE)}
\label{subsec:ALE}

The goal with an ALE plot is to correct the limitation of PDP plots, particularly when the features are correlated. This method was introduced by \cite{ale_pdp}. Just like PDP plots, ALE is a global interpretation method. \\

To understand the interest of the ALE plot, we can take the example of two correlated features like height (called $X_1$) and weight (called $X_2$). If we compute the PDP of the feature $X_1$ at the point $X_1 = 1.80m$ for instance, we will replace the size for all instances by 1.80m even for people weighing 10 kg. So, the PDP computation include unrealistic individuals which leads to a bias in the results.

\begin{figure}[!h]
    \centering
    \includegraphics[width=100mm]{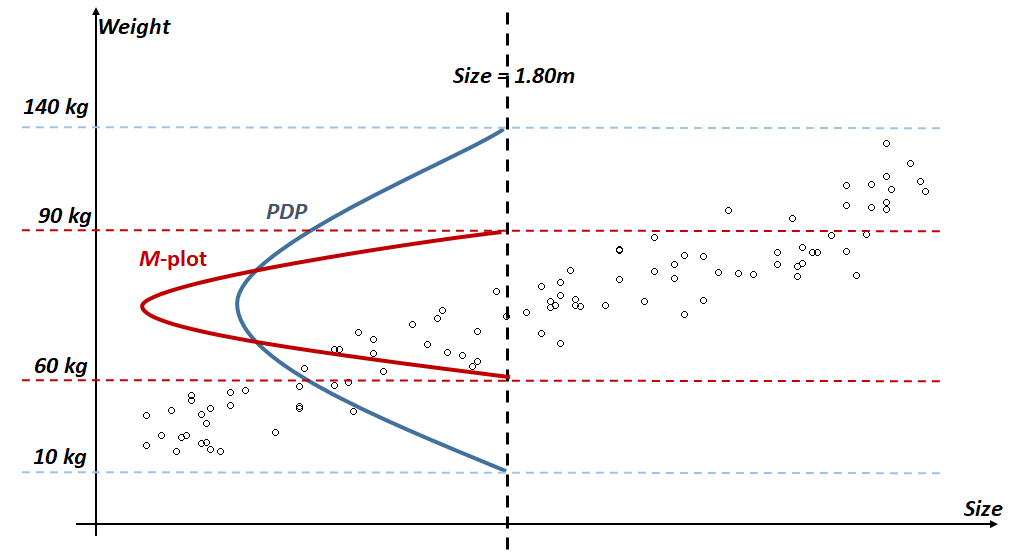}
    \caption{Difference between $M-$ plot which use conditional distribution (red) and PDP which use marginal distribution (blue) }
    \label{fig:LIME}
\end{figure} 

As explained by \cite{molnar2019}, we could consider the marginal plot (called $M-$plot), in which we average over the conditional distribution of the feature. In our example, instead of taking every individual and replacing their size by 1.80m, we consider similar individuals of people measuring 1.80m. So, we will not have unrealistic individuals. 
However, by averaging over all individuals measuring about 1.80m, we compute the combined effect of size and height on the model prediction, because of their correlation.
In the case of our model does not use the feature $X_1$ of the size, but use $X_2$ (the weight), the $M$-plot of the feature $X_1$ wil not be a line as expected but a similar curve to the $M$-plot of $X_2$.

This problem is solved by the ALE plot which is also based on a conditional distribution but makes differences in predictions instead of averages. It allows to isolate the effect of a single feature, instead of all correlated features.
Finally, all these differences are accumulated and centered, which gives the ALE curve. The method for constructing the curve is detailed in Appendix \ref{ALE_methodo}. 

\bigskip

We have just illustrated two methods for analyzing any model in a global way. However, when models are complex or not sufficiently sparse, these tools are sometimes not sufficient to understand the predictions.
The second step of the proposed methodology is the analysis of the interaction between features. 

\subsection{Interaction analysis}
\paragraph{\textbf{Principle of the interaction between variables.}}
The interaction between variables (\textit{feature interaction}) occurs when predictions are not only composed of the sum of the individual effects of each variable, but also of additional terms, corresponding to the fact that the value of one variable also depends on the value of the other variable.
This is the case, for example, when we set up a linear regression model "with interaction":
\begin{itemize}
    \item $Y=\beta_1 X_1 +\beta_2 X_2 + \varepsilon$ is without interaction between $X_1$ and $X_2$
    \item $Y=\beta_1 X_1 +\beta_2 X_2 + \beta_{1,2}X_1 X_2 + \varepsilon$ has an interaction between the explanatory variables $X_1$ and $X_2$.
\end{itemize}

Let us consider another example where we want to predict the average cost of an automobile claim for an policyholder from his age (young or old) and the power of his car (low or high).
We have the  predictions of Table \ref{table:2}.
\begin{table}[!h]
\begin{center}
\scalebox{0.8}{\begin{tabular}{|l|c|r|}
  \hline
  Age & Power & Prediction\\
  \hline
  Young & High & 300 \\
  Young & Low & 200 \\
  Old & High & 250 \\\
  Old & Low & 150\\
  \hline
\end{tabular}}
\caption{Prediction table of Model 1, without interaction}
\label{table:2}
\end{center}
\end{table}

On this very simple model, we can decompose the model prediction in the following way: 
\begin{itemize}
    \item a constant term (\textit{intercept}) of 150
    \item an effect term of the driver's age of 50 (0 if the driver is old, + 50 if he is young)
    \item a vehicle power effect term of 100 (0 if  the power is low, + 100 if it is high).
\end{itemize}
We therefore do not observe any interaction term.
Let us consider another example where the predictions are given in Table \ref{table:1}.
\begin{table}[!h]
\centering
\begin{center}
\scalebox{0.8}{
\begin{tabular}{|l|c|r|}
  \hline
    Age & Power & Prediction\\
  \hline
  Young & High & 400 \\
  Young & Low & 200 \\
  Old & High & 250 \\\
  Old & Low & 150\\
  \hline
\end{tabular}
}
\caption{Prediction table of Model 2, with interaction}
\label{table:1}
\end{center}
\end{table}

In this new model, we can decompose the prediction in this way:
\begin{itemize}
    \item a constant term (\textit{intercept}) of 150
    \item a driver age effect term of 50 (0 if the driver is old, + 50 if he is young)
    \item a vehicle power effect term of 100 (0 if  the power is low, + 100 if it is high)
    \item an interaction term between the age and power variable of 100 (+100 if the insured is both young and owns a powerful car, 0 otherwise). 
\end{itemize}

We have seen the principle of interaction features, we will now present methods to determine the strength of the interaction (with $H$-statistics) and the nature of the interaction (with grouped PDP).

\subsubsection{A method to quantify the strength of an interaction : Friedman's $H$-statistics}
\label{subsec:Hstat}
Using $H$-statistics, we can measure the interaction between variables for any model \cite{H_stat}.
We use the same notations as for Section \ref{subsec:PDP} above on the PDP, namely: $X_S$ represents the subset of variables whose influence we wish to measure, $X_C$ the rest of the variables ($C=\{1,...,n\} \backslash{} S$) and $\hat{f}$ the model, supposedly complex, that we are studying. For any $j\in \{1,...,p\}$, let us denote $PD_j$ the dependency function associated with the variable $X_j$ and $PD_{-j}$ the dependency function associated with all the variables except $X_j$.  Let us also denote, for $j,k\in \{1,...,p\}$ by $PD_{j,k}$ the dependency function associated with the variables $X_j$ and $X_k$.
Recall that we estimate the dependency function using the relation:
\begin{equation}
PD_S(x_S)=\mathds{E}_{X_c}[\hat{f}(x_S,X_C)]\simeq\frac{1}{n}\sum\limits_{i=1}^n{\hat{f}(x_S,x_C^{(i)})}.
\end{equation}

We assume in this section that the model is centered, i.e. 
$\mathds{E}[\hat{f}(X)]=0$.
If there is no interaction between the variables $X_j$ and $X_k$, then we have:
\begin{equation}
    PD_{j,k}(x_j,x_k)=PD_j(x_j)+PD_k(x_k).
\end{equation}
If $X_j$ does not interact with any of the other variables, then the model prediction of an input $x$ satisfies: 
\begin{equation}
    \hat{f}(x)=PD_j(x_j)+PD_{-j}(x_{-j}).
\end{equation}

The coefficients introduced by Friedman exploit this relationship to give a measure of interaction.
The first coefficient, denoted $H_{j,k}$, measures the amount of variance explained by the interaction between $X_j$ and $X_k$:
\begin{equation}
    H^2_{j,k}=\frac{\sum_{i=1}^n\left[PD_{j,k}(x_{j}^{(i)},x_k^{(i)})-PD_j(x_j^{(i)})-PD_k(x_{k}^{(i)})\right]^2}{\sum_{i=1}^n{PD}^2_{j,k}(x_j^{(i)},x_k^{(i)})}
\end{equation}
If there is no interaction between $X_j$ and $X_k$, the $H$-statistic is zero, while if all the variance of $PD_{j,k}$ is explained by the sum of the individual dependency functions, then it is 1.

A second statistic was introduced by Friedman to measure the effect of one variable with all the others:
\begin{equation}
    H^2_{j}=\frac{\sum_{i=1}^n\left[\hat{f}(x^{(i)})-PD_j(x_j^{(i)})-PD_{-j}(x_{-j}^{(i)})\right]^2}{\sum_{i=1}^n\hat{f}^2(x^{(i)})}
\end{equation}

The $H$-statistic is a relatively intuitive measure of interactions, however it is relatively time-consuming to calculate. When the volume of data is large, it can even become impossible to calculate. In such cases, the available data can be subsampled, but this increases the variance of the estimate and makes the $H$-statistic unstable. 

We have seen a method to quantify the strength of the interaction, we will now focus on methods to determine the nature of the interaction.

\subsubsection{Methods to determine the nature of the interaction between features}

Once an interaction is detected, we want to understand how it materializes. For that, we can used the PDP presented before, for different groups. 
By analyzing the PDP differences observed between each group, we can conclude on a combined effect of several features.
Examples of these grouped PDP are presented in the application and is not detailed in this part.

In the first two parts we have methods to understand the role and the importance of each feature, and the nature and the strength of the interactions between them. We will now focus on local interpretation methods which aims to explain the model at an instance level.

\subsection{Local analysis}
This part focus on local analysis, which refers to the most granular interpretation by understanding a specific prediction made for one individual.

We will present the two mains local methods used in practice : LIME and SHAP.

\subsubsection{LIME}
\paragraph{\textbf{Presentation}.}
LIME \cite{lime} is one of the first local approaches that appeared in the field of interpretable machine learning. This method consists in using a substitution model (denoted $M_2$) which locally approaches the model to interpret (denoted $M_1$).\\

First, we apply small perturbations on the initial data $X$ to create a new sample, denoted $\tilde{X}$. We get the predictions of the model to interpret $M_1$ to get the predicted variable for this new sample. Let $M_1$: $\hat{y_1}=f_1(\tilde{X})$. Each observation in the simulated $\tilde{X}$ sample is then weighted according to its proximity to the initial data: the closer it is to the initial data, the greater its weight. On these weighted data, a simple interpretation model $M_2$ is then constructed. This model is generally of Lasso type for regression and a decision tree for classification. Note that this time the model $M_2$ provides a good local approximation but not necessarily a global approximation.\\

The function $\hat{g}$ associated with the local model $M_2$ is trained by solving the following optimization program :  

\begin{equation} 
\hat{g}= \underset{g \in G}{argmin} [J(f,g,\pi_x) + \Omega(g)],
\label{eq:LIME}
\end{equation}

where $J$ is the cost function, $f$ is the function associated with model $M_1$, $g$ the function associated with model $M_2$, in the class of models $G$; $\pi_x$ a measure of proximity that sets the size of the locality around $x$ for the interpretation and $\Omega$ a function to reduce the complexity of the model.\\

However, in practice the implementation in Python of LIME only optimizes the cost function. Hence the user has to choose a model that is not too complex (e.g. if $f_1$ is a regression, a model with a limited number of explanatory variables). \\

\begin{figure}[!h]
    \centering
    \includegraphics[width=100mm]{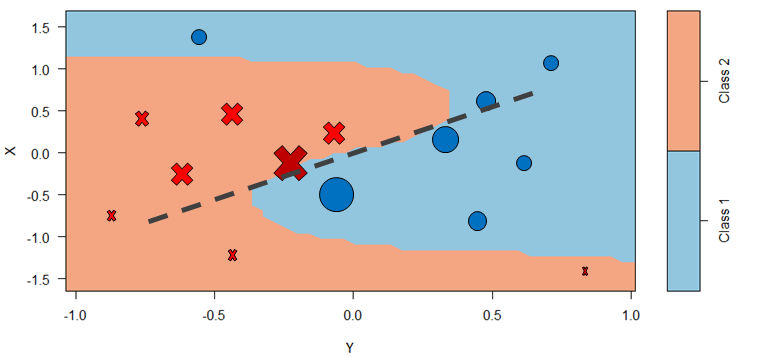}
    \caption{Principle of LIME for a binary classification model \cite{lime}}
    \label{fig:LIME1}
\end{figure}

Figure \ref{fig:LIME1} summarizes how LIME works in the case of a binary classification model (class 1 or 2), with two explanatory variables. The blue zone represents the points associated with Class 1 according to the model and the light red zone represents the points associated with Class 2. The red crosses and the blue dots represent the simulated data for learning the substitution model. The size of the dots and crosses represents the weight of the considered point, according to its distance from the observation of interest, represented by the red cross. The grey dotted line is the decision limit obtained by the LIME algorithm using a linear model.

\paragraph{Limitations}
LIME has been criticized in mostly two ways. 

First of all, as noted in \cite{limite_lime}, the choice of the kernel used in the LIME algorithm to measure the proximity of observations is crucial. It can indeed have a major impact on the accuracy of the resulting explanation. 
\cite{molnar2019} shows an example in which the fitting of the surrogate linear regression is totally unsteady following the chosen kernel width in the LIME methodology.  The Appendix \ref{alternatives_LIME} details an alternative LIME proposed by \cite{BreakDownLive}.

%\begin{figure}[!h]
%    \centering
%    \includegraphics[width=70mm]{pics/lime-fail-1.png}
%    \caption{The importance of the choice of the kernel size to measure proximity in LIME \cite{molnar2019}}
 %   \label{fig:limit_LIME}
%\end{figure}

The other kind of criticism is geared more generally against local interpretation methods. \cite{why_trust_LIME} underline the uncertainties related to the methods of local interpretation of machine learning models, including LIME, and warn users about the robustness and confidence one can have in these methods. Then the difficulty is not so much having a complex model but more how to correctly use the local methods to interpret the model. 
\cite{why_trust_LIME} list a few examples: 
\begin{itemize}
    \item \emph{Randomness introduced by the data sampling}: Because sampling is random, two draws do not necessarily give the same explanation for a prediction.
    \item \emph{The sensitivity of explanations to the choice of parameters}, such as the sample size and the kernel size.
    \item \emph{The variation in the credibility of interpretation according to the studied observation}.
\end{itemize}

\paragraph{\textbf{Conclusion.}}
Even if LIME has some limitations, this method offers a first tool for local interpretation and allows in particular to complement the limitations of standard methods such as the importance of variables in tree-based methods (Random Forest, Gradient Boosting, etc.).

\subsubsection{SHAP}

SHAP \cite{shap} is also a local interpretation algorithm. It is based on the Shapley measure introduced in game theory in 1953.

\paragraph{\textbf{Shapley value in game theory.}}

When a model makes a prediction, we intuitively perceive that each variable does not play the same role: some have almost no impact on the decision made by the model, while others have much more influence. The objective of the SHAP algorithm is to quantify the role of each variable in the final decision of the model. To do this, the algorithm relies on the Shapley value \cite{winter2002shapley} - let us first introduce this value.\\

Let us consider a game $J$ characterized by a 2-uplet : $J=(P,v)$ where $P=(\{1,...,p\} $ is a set of $p$ players, $p \in \mathds{N}^*$ and $v: S(P) \rightarrow \mathds{R}$ is a characteristic function such that : $v(\emptyset)=0$  with $S(P)$ the set of the subsets of $P$.\\

A subset of  players $S \in S(P)$ is called the coalition and the  set of all players $ \{1,...,p\}$ is called the grand coalition. The characteristic function $v$ describes the importance of each coalition.\\

The objective of the game is then to distribute the importance of each player in the total winnings as "fairly" as possible. Thus, we look for an operator $\phi$, which assigns to the game $J=(\{1,...,p\},v)$ a vector $\phi=(\phi_1,...,\phi_p)$ of payoffs. How to define the notion of fair distribution between players? Lloyd Shapley proposed in 1953 a definition in four axioms:

\begin{itemize}
    \item Efficiency:
    $\sum\limits_{i=1}^p{\phi_i(v)}=v(\{1,...,p \})$
    \item Symmetry:
    For any couple of players $(i,j) \in \{1,...,p \}^2$, if $\forall S \in S(\{1,...,p \}\backslash{\{i,j\}})$, $v(S \cup {i})=v(S \cup {j})$, then $\phi_i(v)=\phi_j(v)$ 
    \item Facticity:
    Let $i \in \{1,...,p \} $ a player. If $\forall S \in S(\{1,...,p \}\backslash{\{i\}}), v(S \cup \{i \}) = v(S)$, then: $\phi_i(v)=0$.
    \item Additivity:
    For all games, $v: S(P) \rightarrow \mathds{R}$, $w: S(P) \rightarrow \mathds{R}, \ \phi(v+w)=\phi(v)+\phi(w)$ with: $\forall S \in S(\{1,...p \}), (v+w)(S)=v(S)+w(S)$.
\end{itemize}

The value of Shapley $\phi$ is then the only "fair" value that distributes the total gain $v(\{1,...,p\})$, i.e. the one that meets the four previous conditions. Shapley demonstrates this theorem and gives an explicit value, namely:
\begin{equation} \forall i \in \{1,...,p\}, \ \phi_i(v)=\sum\limits_{S \in S(\{1,...,p\}) \backslash{\{i\}}}{\frac{(p-|S|-1)! |S|!}{p!} (v(S \cup \{i\})-v(S))}.
\label{Shapley}
\end{equation}

\paragraph{\textbf{The value of Shapley applied to model interpretability.}}

SHAP \cite{shap} takes the value of Shapley to make it a measure of the weight of each variable in the predictions of a model, regardless of its complexity. \\

Consider a numerical variable to be predicted, $Y \in \mathds{R}$, from a vector $X \in \mathds{R}^p$ of $p \in \mathds{N}$ explanatory variables.
We suppose that we have a sample $y=(y_1,...,y_n) \in \mathds{R}^n$ corresponding to the target values and $x=(x_{ij})_{1 \leq i \leq n, 1 \leq j \leq p}$ corresponding to the explanatory variables (with $n \in \mathds{N}$ the number of individuals).
Our learning machine algorithm $M$ is calibrated on this sample and we denote by $\hat{g}$ the function associated with the model.\\
%, i.e. the function that returns  $\hat{y}$, the prediction of $y$ made by the model from  $x$: $\hat{y}=\hat{g}(x)$.\\

If we make the analogy with the version of Shapley's measure in game theory, we get:
\begin{itemize}
    \item the game: the prediction task for an instance $\tilde{x} \in \mathds{R}^p$  of the dataset,
    \item the payoff: the current prediction of this instance minus the average prediction of all instances in the dataset,
    \item the players: the values of the characteristics $x_j, \ j \in \{1,...,p\}$, which collaborate to receive the win (here it is a matter of predicting a certain value).
\end{itemize}

Suppose that our variable $Y$ to be explained is the price of a car in \textit{Euros} and that our explanatory variables are $x_1$ and $x_2$, respectively the number of horsepower in the car and the number of doors. Suppose also that for $x_1=150$ and $x_2=4$, the estimated price of the model $\hat{g}$ is $\hat{y}=150,000$. We also know that from the initial data (consisting of several car prices and associated explanatory variables), the mean prediction is $170,000$ euros.
The objective of the game is then to explain this difference of $-20 000$ euros, between the prediction made by the model and the average prediction. For example, we could obtain the following result: $x_1$ contributed $+ 10,000$ euros and $x_2$ for $- 30,000$ euros (compared to the predicted average value) and would therefore justify the observed difference of $- 20,000$ euros.\\

Finally, Shapley value can be defined as the average marginal contribution of an (explanatory) variable across all possible coalitions. 

\paragraph{\textbf{A special case: the value of Shapley in the case of linear regression.}}

We consider a linear model: $\hat{g}(x)=\beta_0 + \sum\limits_{i=1}^{p}{\beta_i \ x_i}$, with $(\beta_i)_{0 \leq i \leq p} \in \mathds{R}^p$.
We then define the Shapley value of the variable $j \in {1,...,p}$ associated with the prediction $\hat{g}(x)$:
$\phi_j(\hat{g})=\beta_j  x_j - \mathds{E}[\beta_j X_j]=\beta_j(x_j - \mathds{E}[X_j])$ (with $\mathds{E}[\beta_j X_j]$ the average effect of the variable $x_j$).
The Shapley value  $\phi_j(\hat{g})$ is also referred to as the contribution of the variable $x_j$ in the prediction of $\hat{g}(x)$ because it is the difference between the effect of the variable and the mean effect.
Note that the sum of the contributions of all the explanatory variables gives the difference between the predicted value for $x$ and the average prediction value. Indeed:

\begin{equation}
    \sum\limits_{j=1}^{p}{\phi_j(\hat{f})}=\sum\limits_{j=1}^p{(\beta_j x_j - \mathds{E}[\beta_j X_j])}= (\beta_0 + \sum\limits_{j=1}^p{\beta_j x_j}) - (\beta_0 +\sum\limits_{j=1}^p{\mathds{E}[\beta_j X_j] })=\hat{g}(x) - \mathds{E}[\hat{g}(X)]
    \label{eq:shap_reg_lin}
\end{equation}
This writing can then be generalized to any model using the Shapley value.

\paragraph{\textbf{The value of Shapley in the general case.}}

Consider a numerical variable to be predicted $Y \in \mathds{R}$, from a vector $X \in \mathds{R}^p$ of $p$ explanatory variables.
We place ourselves in the framework of any model, with $\hat{g}$ the associated function. 
Let $\tilde{x}=(\tilde{x_1},...,\tilde{x_p})$ be the instance for which we want to explain the prediction. \\

Let us define the difference in prediction of a subset of the characteristic values in a particular instance $\tilde{x}$, introduced by \cite{shap_fast}. This is the change in prediction caused by the observation of these values of the explanatory variables.
Formally, let $S=\{i_{1},...,i_{s}\} \subset \{1,...,p\}$ be a subset of the explanatory variables (with $s \in \{1,...,p\}$). Let us note $\Delta^{\tilde{x}}$ the prediction difference, associated with the $S$ subset:
$$\Delta^{\tilde{x}}(S)=\mathds{E}[\hat{g}(X_1,...,X_p)| X_{i_1}=\tilde{x}_{i_1},...,X_{i_s}=\tilde{x}_{i_s}]-\mathds{E}[\hat{g}(X_1,...,X_p)]$$
This difference in prediction corresponds to our cost function.
Thus $(\{1,...,p\}, \Delta^{\tilde{x}})$ forms a coalition game as defined in the previous part.\\

The contribution of the explanatory variable $x_j,\ j \in \{1,...,p\}$, is defined as the Shapley value of this cooperation game $(\{1,...,p\}, \Delta^{\tilde{x}})$.

\paragraph{\textbf{Algorithm for approximate calculation of the Shapley value.}}

The problem, in practice, is the calculation time of the Shapley value due to its complexity (increasing with the number of variables and modalities). Indeed, to do this, we have to calculate all the possible coalitions with or without the variable we want to explain: the complexity is therefore exponential. \\

To remedy this problem, \cite{shap_fast} propose an approximation based on Monte Carlo simulation methods:
$$\hat{\phi_j}=\frac{1}{M}\sum\limits_{m=1}^{M}({\hat{g}(x_{+j}^m)- \hat{g}(x_{-j}^m})),$$
where $j\in \{1,...,p\}$ is the index of the variable we want to explain, $M \in \mathds{N}$ is the number of iterations chosen and $\hat{g}(x_{+j}^m)$ is the prediction for the vector $x=(x_1,. ...,x_p)$ of $p$ explanatory variables, but with a random number of characteristics replaced by a random point, except for the value of the chosen $j$ characteristic. The prediction
$\hat{g}(x_{-j}^m)$ is almost identical to $\hat{g}(x_{+j}^m)$ except that the value $x_j^m$ is also taken from the sample of $x$.\

\paragraph{\textbf{Properties and limits of SHAP.}}

%Although SHAP is also a local interpretation model, it differs from LIME: SHAP explains the difference between a prediction and the global average prediction, while LIME explains the difference between a prediction and a local average prediction.\\

SHAP is one of the few methods of interpretability, to date, with a mathematical basis (see the method Node Harvest \cite{meinshausen} for another example). Indeed, the difference between the prediction and the average prediction is distributed in a fair way between the different variables used by the model, thanks to the efficiency property of the Shapley value. This is not the case with LIME, which is based on a principle that seems coherent but has no mathematical justification. 
SHAP could thus be a method of model interpretability that meets the requirements of the "right to explanation" established by the GDPR. \\

The SHAP method provides an explanation of the prediction made by any model (no matter how complex) by assigning a contribution value to each variable used, unlike LIME which returns a more concise answer, penalizing complex models. We can then consider that SHAP performs fewer approximations than LIME and therefore provides a more precise explanation. \\

When the model to be interpreted is trained with a large number of variables, the interpretation provided by SHAP is not parsimonious. SHAP actually returns as many coefficients as explanatory variables, which sometimes makes it difficult to read.
To get around this problem, an adaptation of SHAP, called \textit{Kernel Shap} (Linear LIME + Shapley Values) is proposed \cite{shap}. The idea is to link the Equations \ref{eq:LIME}  (from LIME) and \ref{Shapley} (from SHAP). By judiciously choosing the cost function $J$, the proximity measure $\Pi_{x'}$ and the regularization term $\Omega$, it is then possible to write the Shapley value as the solution to the optimization problem posed by LIME in Equation \ref{eq:LIME}. This combination then makes it possible to provide more parsimonious explanations. \\

Moreover, SHAP in its initial version assumes that the variables are independent. Nevertheless, an alternative has recently been proposed by \cite{SHAP_dependant}. \\

Finally, note that SHAP provides only an indication of the contribution of each variable for a given prediction. It does not allow the inference of overall effects, contrary to the interpretation of \textit{odds-ratios} in the linear regression framework. It only provides a local understanding, even if the latter is sometimes more explicit when using complex models such as neural networks or ensemble methods (random forests, XGBoost for example).\\

In the last two parts, we first explained the importance of the need for interpretability in the use of predictive models, while giving some definitional elements to this difficult-to-define concept, and then we presented some common methods of interpretability. 
To conclude this section, we would like to emphasize the need to remain cautious and critical in the use of the various interpretability methods.
The reliability and robustness of interpretability methods is an important issue, especially in the case of local methods, which can lead to erroneous interpretations.  \cite{Alvarez} insists on the robustness of local methods, such as LIME and SHAP, and shows on a few examples, that close entries can lead to significantly different interpretations with LIME or SHAP. It is possible to quantify the robustness of a local interpretability method by measuring a local Lipschitz constant around an observation. This type of approach is interesting and it is even necessary before drawing definitive conclusions on a particular example to make sure that the results provided by a particular interpretability method are robust with respect to the input variables. It is natural to expect that interpretability methods are as robust as the models being evaluated.  We can also mention another approach, proposed in \cite{Alvarez1}, where interpretability is taken into account in the construction phase of  a neural network, in order to obtain a more robust explainability.
More generally, the reader may wish to consult the papers \cite{Mohseni} and \cite{Doshi} on the evaluation of interpretability methods.\\

Let us now apply these methods to a concrete case in order to illustrate their contributions and their limits.

\section{An application of interpretation methods to automobile insurance pricing to show how interpretation tools can help model users according to the audience} 

In this section, we apply the interpretation methods to an actuarial case study, automobile pricing. We want to understand whether more complex models, possibly more efficient, could be interpreted in the same way as the more traditional tools used today, and more specifically, whether these "black box" models could really be implemented while complying with the regulations (right to explanation provided for by the GDPR, controls of prudential authorities, etc.).

\subsection{Modelling and comparison of the different approaches}
\begin{figure}[!h]
    \centering
    \includegraphics[scale = 0.7]{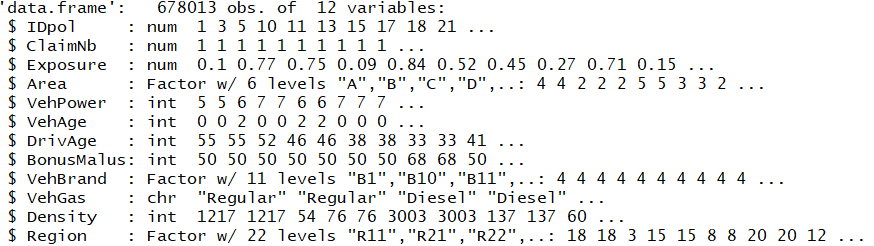}
    \caption{Description of the $freMTPL2freq$ data, used here to model the frequency of claims}
    \label{fig:description_data}
\end{figure}

We use the public database \textit{freMTPL2freq}, available in the R $CASdatasets$ package. This is data from a French motor liability insurance portfolio for different policyholders observed over one year. This database allows to model the frequency of claims. We have more than 600,000 insurance policies, with explanatory variables such as the age of the policyholder, the vehicle's power or its age. The particularity of the frequency data is that it is very unbalanced, with many zeros (no accident) and exposures that vary greatly. A more detailed description of the data is provided in Figure \ref{fig:description_data}. 

For the pricing of car insurance, the classic actuarial approach is an independent modelling of frequency (number of annual claims) and severity (average cost of a claim), to form the pure premium by combining the predictions of these two models. We focus solely on the frequency part here. On the severity part, it is to be noted that the modeling is based on a small data set, since only the policyholder who have sustained a loss are used. Moreover, the correlation between the explanatory variables and the target variable (claim amount) is generally quite low, which makes it difficult to obtain a good predictive model. We observed during our study that it is difficult to significantly improve the performance of the classical GLM Gamma, even with complex models such as ensemble methods. Let us now focus on the frequency model.

Before the fitting of the algorithms to address this problem, we performed preliminary treatments to replicate operational practices: analysis of outliers and extreme claims (which were capped), processing of categorical variables, etc. (cf \cite{Delcaillau} for more details). This last step is essential for the implementation of a Generalized Linear Model (GLM) because, without pre-processing of the data, monotony is imposed on the numerical variables due to the linear nature of the model. We have taken up the pre-processing proposed in the article \cite{case_study}. 

Our objective in this section is to compare the interpretability of two models: a classical GLM model, very often used in actuarial science, and another model, more complex, that shows better performances. We train those models and then use the interpretation tools presented above to show that it is possible to understand the "black box" model. 

In the range of complex models such as Random Forest or neural networks, we finally opted for an eXtreme Gradient Boosting (XGBoost) model \cite{xgboost}. XGBoost has become very popular in many machine learning competitions, such as Kaggle, thanks to the flexibility provided by its numerous hyperparameters. We have optimized them based on cross validations. Throughout this section, we denote by $A$ the trivial model, returning the average frequency of claims, $B$ the best GLM (Poisson) model found and $C$ the qualified black box model, which is an XGBoost.\\

\begin{figure}[!h]
    \centering
    \includegraphics[scale = 1]{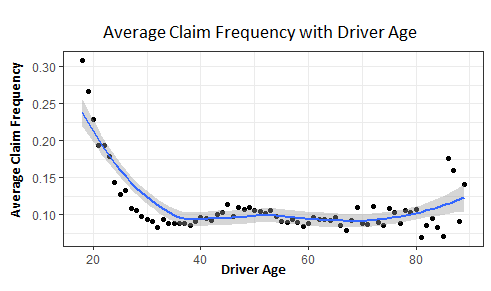}
    \caption{Empirical relation observed in the data between the average frequency of claims and the driver's age}
    \label{fig:courbe_en_u}
\end{figure}

It is important to note that in the case of the GLM, it is imperative that the numerical variables are pre-processed beforehand and made categorical. Indeed, without it, the relationships between the numerical explanatory variables and the output would all be monotone. In particular, the classically observed "U" curve representing the relationship between the driver's age and the average frequency of claims could not be obtained without this pre-processing (cf. Figure \ref{fig:courbe_en_u}). In the context of XGBoost, and more generally non-linear models, this pre-processing is not necessary, as these models are capable of capturing these complex relationships. To be able to compare the GLM and the XGBoost, we completed the analysis by fitting a new XGBoost model, noted \textit{C-cat}, which uses the same variable transformation as the GLM. The parameters of this model were also optimized using cross-validation. 
 It is common in the insurance industry that actuaries request to keep some binned variable. We want to emphasize some situations that are commonly faced when it comes to challenge traditional GLM with other models. Classes are often a mix of expert judgement and model outcomes. Moreover, as the paper does not focus on the modeling itself but on the interpretation, we decided to keep an ISO format of the data to focus on model comparison/interpretation. As interpretation tools could help to get a similar level of model transparency, a further step would be to use the full potential of complex models and challenge the prior introduced in the data (by for instance removing categorized variables).\\

The results for these different models are given in Table \ref{tab:res_glm_xgb}. The evaluation metric we use here is Poisson deviance, which is very commonly used for frequency data. We also give the Mean Squared Error (MSE) and Mean Absolute Error (MAE) values as a complementary indicator. We show the relative gains over the trivial model in brackets. 
These criteria were calculated on both the training data (\textit{In-Sample}), and on the test data (\textit{Out-of-Sample}) to test the ability of the model to fit to new data. There is a gain of about 3\% on the Poisson deviance loss of the XGBoost on the best GLM model. In a strongly competitive market, such as automobile insurance, this gain in accuracy could prove essential, especially to avoid bad risks and thus anti-selection. It should be noted, however, that these comments should be qualified since it has not been proven that strong segmentation is synonymous with profit in a competitive market - indeed, segmentation, in addition to being in opposition to the basic principle of insurance, i.e. mutualisation, leads to an increase in the volatility of earnings \cite{segmentationPlanchet}.

% Please add the following required packages to your document preamble:
% \usepackage{multirow}
\begin{table}[!h]
\scalebox{0.88}{
\begin{tabular}{c|c|c|c|c|c|c|}
\cline{2-7}
\multirow{2}{*}{}                        & \multicolumn{2}{c|}{Poisson deviance}                                                                                & \multicolumn{2}{c|}{MSE}                                                                                                & \multicolumn{2}{c|}{MAE}                                                                                                \\ \cline{2-7} % Added multirow component 26.08.2022 Dina
                                         & Train                                                       & Test                                                       & Train                                                       & Test                                                       & Train                                                       & Test                                                       \\ \hline
\multicolumn{1}{|c|}{Baseline ($A$)} & 32.94                                                      & 33.86                                                      & 0.0564                                                     & 0.0596                                                     & 0.0995                                                     & 0.1015                                                     \\ \hline
\multicolumn{1}{|c|}{Best GLM ($B$)}   & \begin{tabular}[c]{@{}c@{}}31.27\\ \textit{(+5.06\%)}\end{tabular}  & \begin{tabular}[c]{@{}c@{}}32.17\\ \textit{(+4.99\%)}\end{tabular}  & \begin{tabular}[c]{@{}c@{}}0.0557\\ \textit{(+1.28\%)}\end{tabular} & \begin{tabular}[c]{@{}c@{}}0.0589\\ \textit{(+1.66\%)}\end{tabular} & \begin{tabular}[c]{@{}c@{}}0.0979\\ \textit{(+1.62\%)}\end{tabular} & \begin{tabular}[c]{@{}c@{}}0.0999\\ \textit{(+1.60\%)}\end{tabular} \\ \hline
\multicolumn{1}{|c|}{XGBoost ($C$)}        & \begin{tabular}[c]{@{}c@{}}30.22\\ \textit{(+8.24\%)}\end{tabular} & \begin{tabular}[c]{@{}c@{}}31.29\\ \textit{(+7.59\%)}\end{tabular} & \begin{tabular}[c]{@{}c@{}}0.0548\\ \textit{(+2.95\%)}\end{tabular} & \begin{tabular}[c]{@{}c@{}}0.0582\\ \textit{(+2.35\%)}\end{tabular} & \begin{tabular}[c]{@{}c@{}}0.0965\\ \textit{(+3.02\%)}\end{tabular} & \begin{tabular}[c]{@{}c@{}}0.0988\\ \textit{(+2.74\%)}\end{tabular} \\ \hline
\multicolumn{1}{|c|}{XGBoost cat (\textit{C-Cat})}        & \begin{tabular}[c]{@{}c@{}}30.34\\ \textit{(+7.89\%)}\end{tabular} & \begin{tabular}[c]{@{}c@{}}31.37\\ \textit{(+7.36\%)}\end{tabular} & \begin{tabular}[c]{@{}c@{}}0.0549\\ \textit{(+2.71\%)}\end{tabular} & \begin{tabular}[c]{@{}c@{}}0.0582\\ \textit{(+2.23\%)}\end{tabular} & \begin{tabular}[c]{@{}c@{}}0.0966\\ \textit{(+2.87\%)}\end{tabular} & \begin{tabular}[c]{@{}c@{}}0.0988\\ \textit{(+2.68\%)}\end{tabular} \\ \hline
\end{tabular}}
\caption{Performances of the different models for the frequency - baseline model, GLM, XGBoost, XGBoost cat}
\label{tab:res_glm_xgb}
\end{table}

However, more than the impact of this gain in accuracy on the insurer's profitability, the article chooses to focus on the ability to interpret the predictions made by this complex model and to understand it the same way as the GLM.

\subsection{Interpretation of the GLM and XGBoost models}
\subsubsection{The GLM model}

\begin{figure}[!h]
    \centering
    \includegraphics[scale = 0.76]{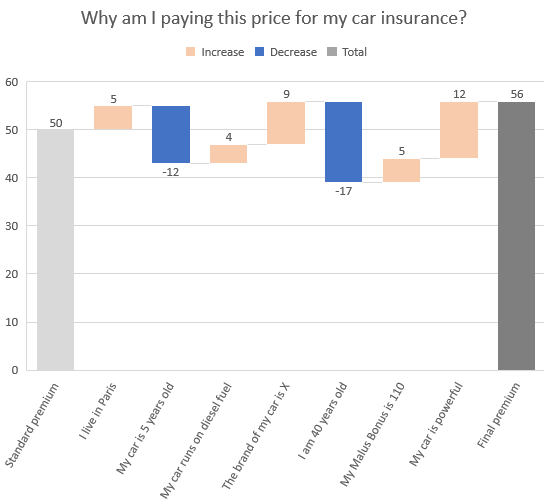}
    \caption{An example of a way to justify the price of a car insurance in regards to the car-owner's characteristics}
    \label{fig:prix_glm}
\end{figure}

\paragraph{\textbf{An intrinsically interpretable model.}}
We started by studying the performance of the two models, and in particular we checked the stability with respect to the sampling of the training and test datasets, and then we put into practice the interpretation methods developed in Section \ref{sec:1_methodesInterpretation}, to better understand the predictions. We first analyzed the predictions of the GLM model. Because GLM are both sparse and simulable (cf Section \ref{sec:0_interpretabilite}), we were able to directly understand the model based on the coefficients of each variable. In particular, we decomposed the path leading to a prediction for a given policyholder (cf Figure \ref{fig:prix_glm}).
It is also easy to study how the prediction of the GLM changes when a policyholder's characteristics are modified: one just needs to look at the change in the coefficient induced by this characteristic modification.

Thus the properties of the GLM help us understand the model predictions, and that is why we can consider it is an intrinsically interpretable model (model-based interpretability, cf Section \ref{sec:0_interpretabilite}).

Because the GLM is intrinsically interpretable and we understand it, we test the relevance of post-hoc interpretability tools we presented in Section \ref{sec:1_methodesInterpretation}: if they work perfectly, we should have the same conclusions based on them than we have with model-based interpretability.

\begin{figure}[!h]
    \centering
    \includegraphics[scale = 0.8]{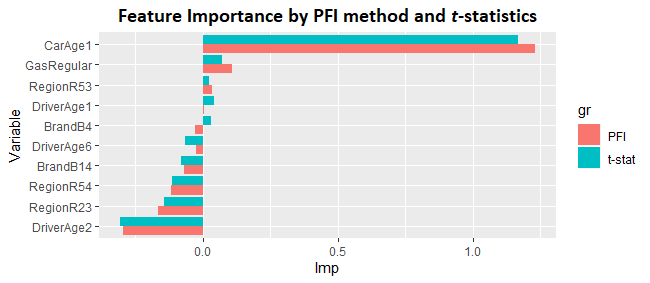}
    \caption{Comparison of feature importance as measured by the $t$-statistic and permutation feature importance ($PFI$).}
    \label{fig:PFI_vs_TSTAT}
\end{figure}

\paragraph{\textbf{Global interpretation.} }
Let us first of all have a look at feature importance. As we have seen in Section \ref{sec:1_methodesInterpretation}, permutation feature importance is a possible approach for this. We want to check that the obtained results are similar to those of $t$-statistics, which is specific to GLM models and is based on a totally different principle.
Figure \ref{fig:PFI_vs_TSTAT} shows the results obtained by these two approaches, which are quite similar. Note that we represented the importance of only a few variables for the sake of readability.

\begin{figure}[!h]
    \centering
    \includegraphics[scale = 0.4]{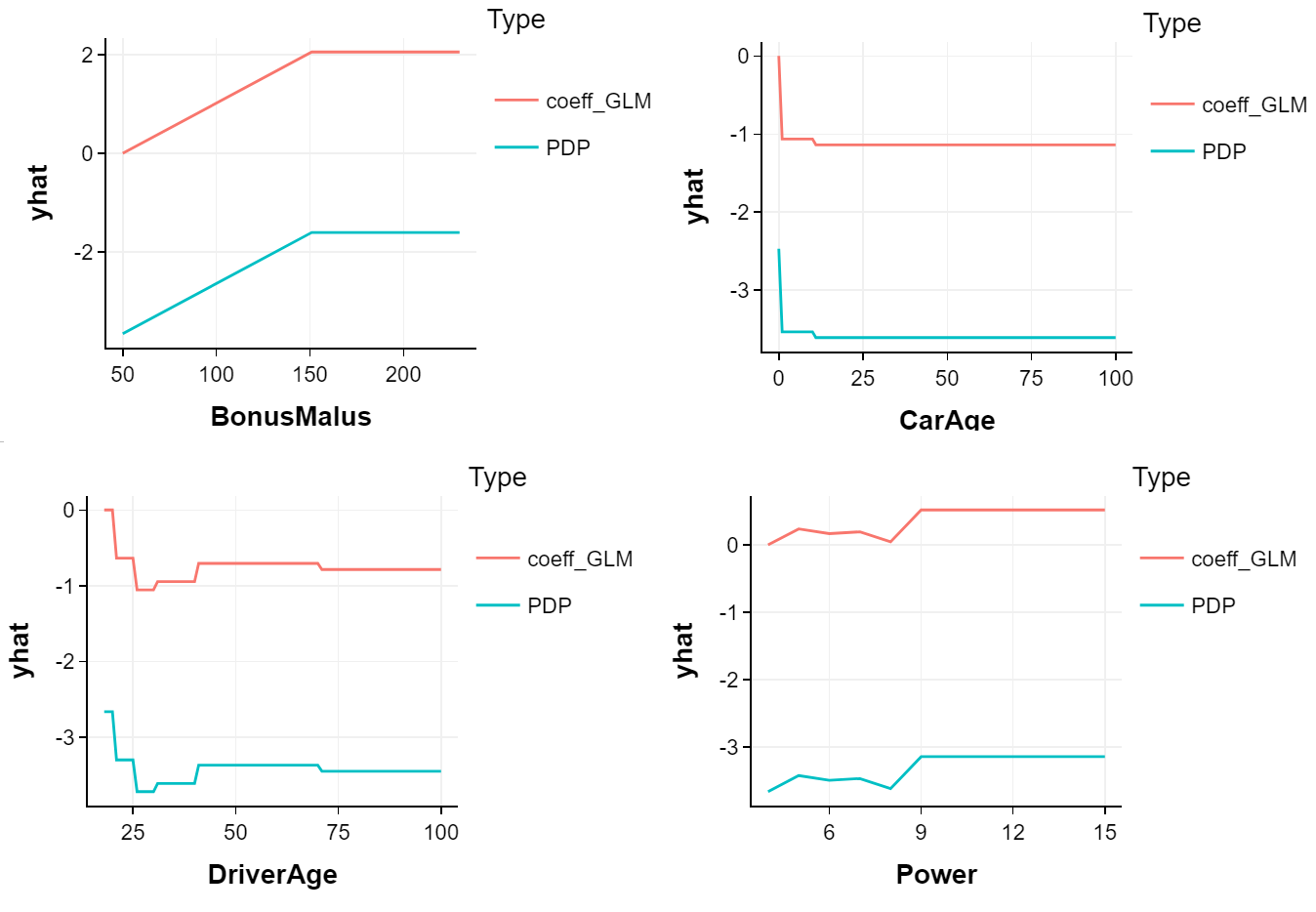}
    \caption{Partial dependence plots (blue) and GLM coefficients (red) of the model $B$ (GLM frequency)}
    \label{fig:pdp_glm}
\end{figure}

We can understand the overall role of each variable in the prediction of the GLM model with feature importance but other interpretability tools give some complimentary insights, like the average impact of each modality on the prediction with Partial Dependence Plots (PDP). PDP measures the average marginal effect of a variable on the prediction (see Section \ref{subsec:PDP}).
As we can see in Figure \ref{fig:pdp_glm}, the PDP curves are simple translations of the different GLM coefficients (up to the inverse link function, here exponential). This is consistent with GLM properties and PDP definitions and it supports the idea that the information of the model coefficients alone is sufficient for its interpretation in the case of a GLM model. Note that we could also use ALE here and it would lead to the similar interpretations as the PDP in the case of the GLM. %TODO: check le commentaire sur la VF

\paragraph{Interactions Analysis}.
As the model we considered is a GLM, there is no interaction between the variables. Note that we could have used a GAM (\textit{Generalized Additive Model}) to include interactions between some variables.

Let us check that the tools of the Section \ref{sec:1_methodesInterpretation} to identify interactions are in adequacy with this analysis. 
First of all, the $H$-statistics (cf. Section \ref{subsec:Hstat}), provide coefficients close to 0 for each variable, which means the absence of interaction within the model. 
%Ceci est bien vérifié sur la figure \ref{fig:H_stat_glm}. 

\begin{figure}[!h]
    \centering
    \includegraphics[scale = 0.42]{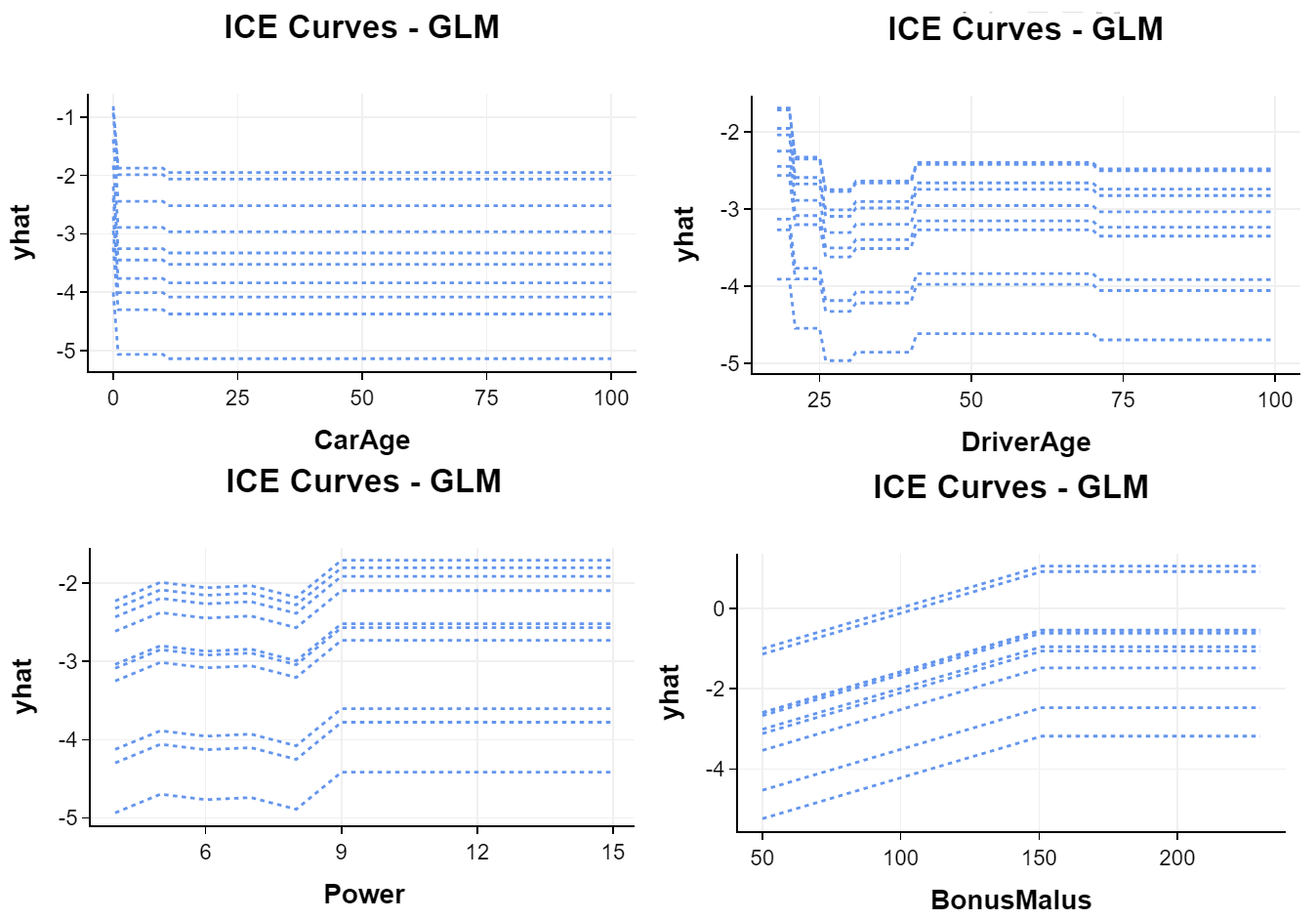}
    \caption{Some ICE curves (with the exception of the inverse link function) associated with 4 variables used in the GLM model}
    \label{fig:ice_glm}
\end{figure}

We can also represent the \textit{ICE} curves which allow us to identify possible interactions between variables. We observe, in Figure \ref{fig:ice_glm}, that these different curves are translated between them (and with the PDP curve which is only the mean), a sign of absence of interaction.

\paragraph{Local interpretation. }
Local interpretation methods (such as \textit{LIME} and \textit{SHAP}) are also coherent with the interpretations we can draw from the coefficients of the GLM. Note that the different values we obtained to interpret a prediction locally can be directly deduced from the coefficients of the GLM model: in fact due to the linearity of the GLM model, the global analysis is self-sufficient.

\begin{figure}[!h]
    \centering
    \includegraphics[scale = 0.65]{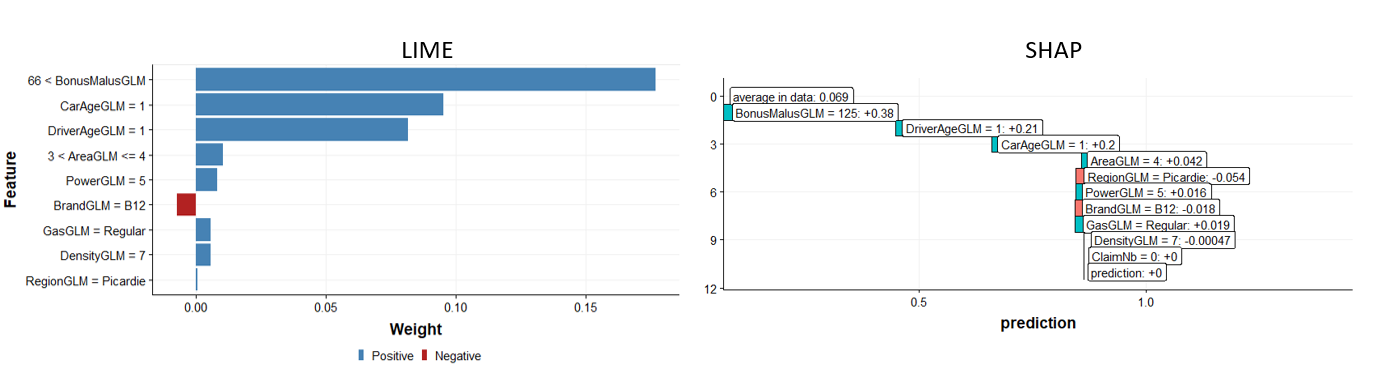}
    \caption{Illustration of results obtained using the local LIME and SHAP interpretation methods on a policyholder for the GLM model $A$}
    \label{fig:lime_shap_glm_diff}
\end{figure}

As we fully understand the GLM model, we can use this example to illustrate the differences between LIME and SHAP (cf Figure \ref{fig:lime_shap_glm_diff}).

If we look at the different contributions of the variables according to SHAP or LIME in Figure \ref{fig:lime_shap_glm_diff}, we note that features contribution look overall similar but are not exactly the same. It is explained by the different approaches of local interpretations and emphasis the fact that LIME or SHAP methodologies (even if they are commonly used) are not representing the ground truth. Both however help to complete global analysis with local results and could confirm expert opinion (or trigger a deeper investigation on the explained model or underlying data). \\

To conclude, we do not really need the methods we introduced in Section \ref{sec:1_methodesInterpretation} to interpret the GLM model results: whether we want to get a global or local understanding, we only need the coefficients of the model and our knowledge of its properties (eg: lack of interactions between variables).

\subsubsection{The XGBoost model}
In the previous section, we have shown that the conclusions we can draw from the different post-hoc interpretation methods were similar to what we were expecting given our knowledge of the GLM model as it is intrisincally interpretable. Now let us use these methods on a model that is often considered as a black box, a XGBoost model. We will go through the same steps: global interpretation, interactions and local interpretation.

As mentioned in the introduction, the interpretability depends on the target audience. In this part, we will investigate the relevance of each method according to the stakeholders involved. We will focus on two types of audience:
\begin{itemize}
\item a policyholder who wants to understand its tariff. The policyholder in question has the characteristics described in Table \ref{tab:policyholder}. The fitted model GLM and XGBoost predict the claim frequency (over one year) respectively of 86.3\% and 94\%; the two values are both greater than the 99\% quantile of the prediction made by each model on the learning database.

\begin{table}[H]
\scalebox{0.88}{
\begin{tabular}{|l|l|l|l|l|l|l|l|l|l|}
\hline
\textbf{Feature}        & Power      & CarAge     & DriverAge   & BonusMalus   & Brand        & Gas              & Area       & Density       & Region            \\ \hline
\textit{\textbf{Value}} & \textit{5} & \textit{0} & \textit{19} & \textit{125} & \textit{B12} & \textit{Regular} & \textit{4} & \textit{1196} & \textit{Picardie} \\ \hline
\end{tabular} }
\caption{Studied policyholder's characteristics}
\label{tab:policyholder}
\end{table}

\item an actuary who wants to understand the fitted model and to check its reliability
\end{itemize}
\paragraph{Global interpretation. }

First, we use the Permutation Feature Importance (PFI) to measure the importance of features. Note that there are also specific methods for XGBoost models, but we will not use them in this paper. 

\begin{figure}[!h]
    \centering
    \includegraphics[scale = 0.95]{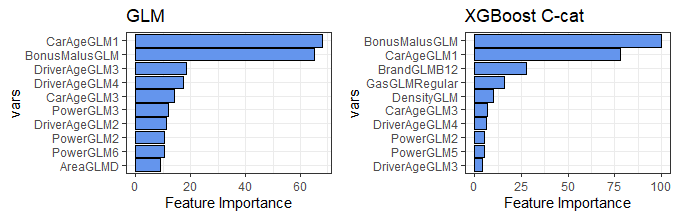}
    \caption{The 10 most important features in the GLM (\textit{A}) and XGBoost (\textit{C-cat}) models}
    \label{fig:imp_var_xgb_glm}
\end{figure}

Figure \ref{fig:imp_var_xgb_glm} gives the importance of the different variables we used (in the XGBoost and GLM). In order to compare the GLM and the XGBoost models, we only analyze the XGBoost model with the pre-processed variables (namely the \textit{C-cat} model) for the feature importance: indeed PFI gives a score for each feature, so we cannot directly compare the importance of the GLM and the \textit{C} XGBoost.

We observe that for the GLM and the XGBoost, the  modality $1$ of the variable $CarAge$ (i.e. car less than one year old) seems to have a preponderant role on the predictions made by these two models, and the same goes for the bonus-malus numerical variable. However, because the PFI method is based on the modalities of each variable, there are 52 coefficients with the preprocessed categorical variables, which makes the analysis difficult and not parsimonious.

Instead we use the $R$ \textit{DALEX} package: that package adapts the PFI method to give only one score to each variable (instead of one by modality). 

\begin{figure}[!h]
    \centering
    \includegraphics[scale = 0.95]{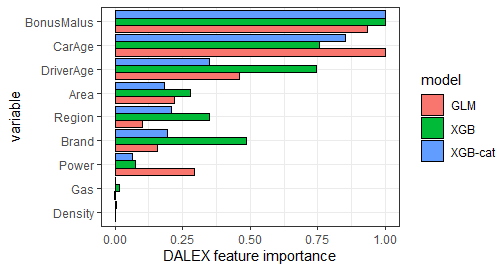}
    \caption{Importance of the variables as measured by permutation with \textit{DALEX}}
    \label{fig:imp_var_dalex}
\end{figure}

Using this method, we have a simplified analysis of our different variables and their roles within the model. It also allows us to compare our three models $B$, $C$ and \textit{C-cat}. Figure \ref{fig:imp_var_dalex} summarizes the importance of the variables of the three models mentioned above: overall the role of the variables is similar within each of the models. In particular, bonus-malus and driver age are preponderant for all three models. We also note that driver age plays a more important role in the XGBoost ($C$) than in the other models. Finally, the variables $Gas$ and $Density$ have very little influence on average on the models prediction. This leads to different local interpretation for the XGBoost models than it does for the GLM. 

Indeed,  we saw previously that in the case of the GLM, for all predictions, regardless of the individual concerned, these variables have no (or very little) impact. That is because the sole knowledge of the coefficients of the GLM model is enough to understand the model's local behaviour: locally the coefficients are applied the same way for all individuals predictions. Hence local methods are not very relevant for the GLM model. On the contrary, for the XGBoost model it means for most individual predictions that these variables have a very limited importance, but it is possible that for some $Gas$ and $Density$ are extremely important locally. This could be checked with local methods such as LIME and SHAP.

Now that we know how important each variable is globally, let us have a look at the average effect of the value of a variable on the predictions, with the PDP and ALE methods. 

One of the advantages of this is that we can check whether we find back the same relations we had empirically found in the dataset in the impact of each variable on the predicted claims. 

\begin{figure}[!h]
    \centering
    \includegraphics[scale = 0.63]{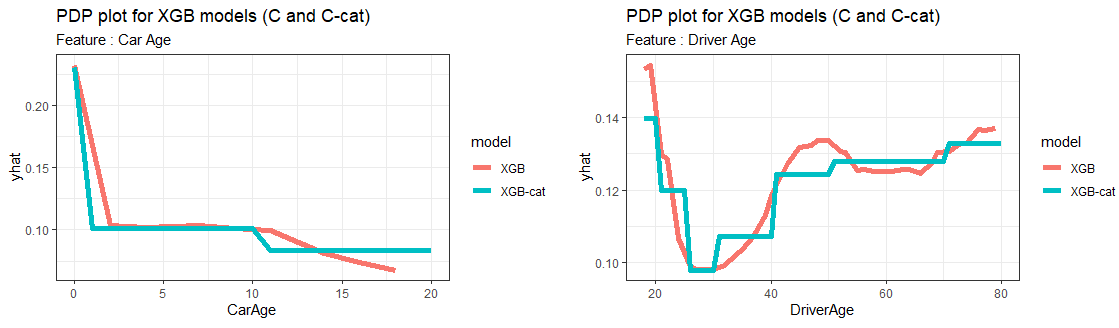}
    \caption{Partial dependence plots of the age of the driver and the age of the vehicle for the $C$ and \textit{C-cat} models}
    \label{fig:pdp_driver_car_age_xgb_xgb_cat}
\end{figure}

In particular, we hope to find a "U" curve for the driver age variable, i.e. a lot of claims for young (under 25) and old (over 70) drivers, and relatively low or moderate claims for the intermediate ages. When we look at the PDP for that variable (Figure \ref{fig:pdp_driver_car_age_xgb_xgb_cat}), we can confirm the model replicates that relation. Note the blue curve has some steps since we discretized the age variable in several age groups ($[18,21[, [21, 26[, [26,31[, [31,41[, [41, 51[, [51, 71[, [71, +\infty[$).

We can carry the same kind of analysis on the age of the vehicle. On the left of Figure \ref{fig:pdp_driver_car_age_xgb_xgb_cat}, we observe that the older the vehicle, the higher the frequency of predicted claims: in other words, the relationship is decreasing. Note that these curves only represent less than 20-year old vehicles. This interpretation is in line with the GLM model. With an empirical study on vehicle age and claim frequency, we can check the relevance of both models. In the database, 30\% of claims indeed occur for vehicles that are less than two years old. One explanation for this can be the strong correlation between older drivers, who are more exposed to claims, and the fact that they change vehicles more regularly. However, this cannot be confirmed with current data. 

You can find the ALE plot for all of the other variables in the appendix (Figure \ref{fig:pdp_all_xgb_freq_num}). Note for instance the increasing relation between the target variable and the bonus-malus.\\

\textbf{Relevance according to the target audience} \\
Feature Importance and PDP are important tools for an actuary to understand the global behaviour of the model and the relevance with business expertise. The fact that the three main features are Bonus Malus, DriverAGe and CarAge is coherent, just like the monotonicity of the prediction with the CarAge (decreasing curve) and the BonusMalus (increasing curve).  \\
For a policyholder, these methods are less important because they do not allow to understand exactly the price paid. However, it can be an interesting knowledge for the policyholder to know which feature contribute the most in the paid price (Feature Importance) and with which average effect on the prediction (PDP/ ALE). For the specific example we took, the policyholder could easily understand the high probability of claims predicted by the model with these two methods. In fact, the three most important features (Bonus Malus, DriverAge, CarAge) are all in the range where PDP/ALE are maximal. However, heterogeneous effects are not considered and these tools are not sufficient to explain precisely the prediction.

\paragraph{\textbf{Interactions.}}
So far, we have only studied univariate effects with the PDP plots, let us now analyze multivariate effects. An XGBoost model allows for more interactions than a GLM, so it can be more complex to explain: the prediction of an XGBoost model indeed cannot be decomposed as the sum of the effect of each variable. But as \cite{trees_interac} explain, tree-based methods like XGBoost are particularly good at modeling interactions between variables. 

A first solution to show model interactions is the H-statistic. This statistic, that we presented in Section \ref{subsec:Hstat}, captures the strength of the interaction by measuring the proportion of the variance that is due to the interaction between several variables. It is mostly computed using partial dependencies, comparing the variance due to the interaction and the total variance. The value of the H-statistic is between 0 and 1, with 0 referring to the absence of interaction and 1 indicating that the prediction is purely guided by the interaction studied.

In this example, we only consider two variables at a time. As the H-statistics formula is based on partial dependencies, the computation time is very high and we have to make approximations.

\begin{table}[ht]
\scalebox{0.8}{
\centering
\begin{tabular}{rrrrrrrrrr}
  \hline
 & Area & BonusMalus & Density & Region & Power & DriverAge & Brand & Gas & CarAge \\ 
  \hline
Area &    & 0.2\% & 0.2 \% & 0.1\% & 0.2\% & 0.1\% & 0.1\% & 0.3\% & 0.6\% \\ 
  BonusMalus &    &    & 1.4\% & 1.5\% & 1.1\% & 1.4\% & 1.6\% & 2.9\% & 4.5\% \\ 
  Density &    &    &    & 1.3\% & 1.1\% & 0.6\% & 0.7\% & 0.9\% & 6.3\% \\ 
  Region &    &    &    &    & 0.6\% & 0.5\% & 0.4\% & 1.1\% & 4.3\% \\ 
  Power &    &    &    &    &    & 0.5\% & 0.5\% & 1.5\% & 11\% \\ 
  DriverAge &    &    &    &    &    &    & 0.5\% & 1.1\% & 2.1\% \\ 
  Brand &    &    &    &    &    &    &    & 1.3\% & 2.7\% \\ 
  Gas &    &    &    &    &    &    &    &    & 6.7\% \\ 
  CarAge &    &    &    &    &    &    &    &    &    \\ 
   \hline
\end{tabular}
}
\caption{$H$-statistics for the different combinations of the XGBoost variables (on 10,000 points, that led to a long computation time)} 
\end{table}

However, in this example, we only use categorical variables, which makes the H-statistics irrelevant. The $H$-statistic indeed overestimates the effect of interactions with categorical variables, like the combination $(Power, Gas)$ (respectively 6 and 2 modalities). 

Another drawback of the H-statistic is that it does not describe how the interaction of two variables changes the prediction (only its intensity). \emph{Individual Condition Expectation} plots (ICE) on the contrary give this kind of information. ICE generalize on partial dependencies for each observation (cf. Section \ref{subsec:PDP}), which makes it a method for local interpretation. Analyzing interactions with ICE plots is straightforward: if all curves are not simple translations of one another, it means there are interactions between variables. 

\begin{figure}[!h]
\centering
\begin{tabular}{ccc}
\includegraphics[scale = 0.45]{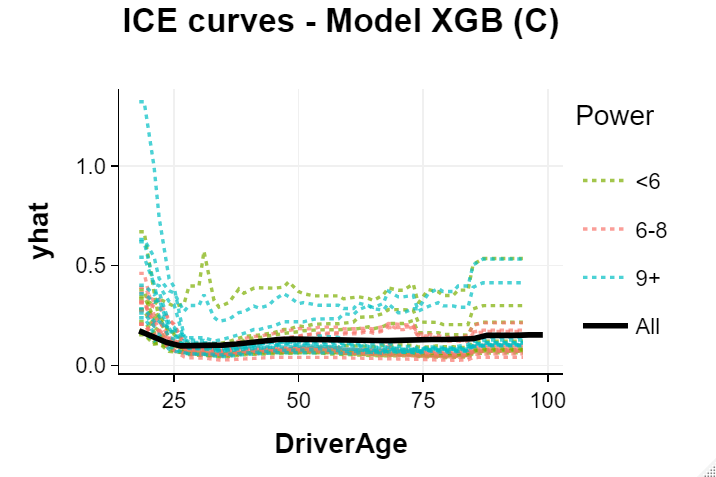} &

\includegraphics[scale = 0.23]{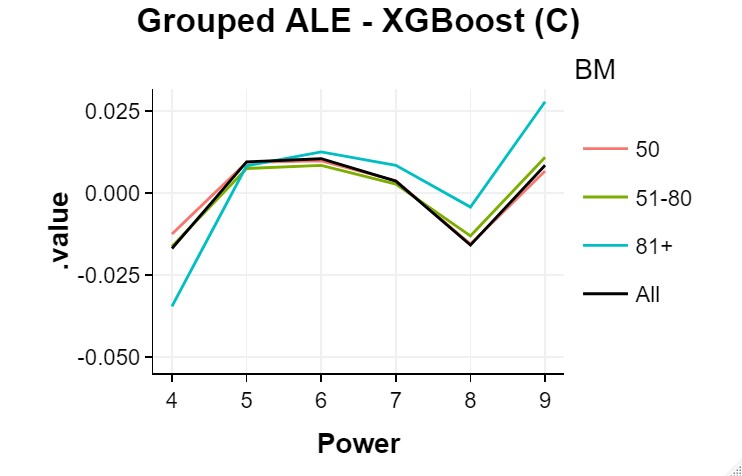}
\end{tabular}
\caption{On the left: ICE plots of the $C$ XGBoost model for the driver's age and the vehicle power. On the right: Grouped ALE plots of the $C$ XGBoost model for the bonus-malus and the vehicle power}
    \label{fig:ice_groupedALE}
\end{figure}

By the definition of an XGBoost model, we expect interactions - i.e. ICE curves that are not translated. To highlight more precisely the interactions, we put in a different color each curve depending on the modality taken by a chosen variable. We have chosen to study the effect of the age of the driver and the power of the vehicle, divided here into 3 classes. Several studies show that the joint effect of a young driver and a high-powered vehicle drastically increased the risk of claims (in terms of frequency). This can be observed on the left of Figure \ref{fig:ice_groupedALE}: the blue curves represent policies with the highest vehicle power and the black partial dependence curve represents the average effect of the driver's age on the prediction. For blue curves, the slope tends to be bigger for young ages (between 18 and 25) than the PDP, especially for one ICE curve, which indicates a potential additional risk of claims when the policyholder is young and has a powerful car.

%\begin{figure}[!h]
%    \centering
%    \includegraphics[scale = 0.65]{pics/ice_driver_age_power2.png}
%    \caption{ICE plots of the $C$ XGBoost model for the driver's age and the vehicle power}
%    \label{fig:ice_driver_age_power}
%\end{figure}

On the right Figure \ref{fig:ice_groupedALE}, you can see the joint effect of bonus-malus and vehicle power, where the slope between power 8 and 9 is higher for large Bonus Malus (class 81+ in blue) compared to other Bonus Malus classes (in red and green) and the PDP (in black).

%\begin{figure}[!h]
%    \centering
%    \includegraphics[scale = 0.28]{pics/ale_power_BM2.jpg}
%    \caption{Grouped ALE plots of the $C$ XGBoost model for the bonus-malus and the vehicle power}
%    \label{fig:ice_BM_power}
%\end{figure}

Note also that the ICE curves highlight atypical individuals, for whom the number of predicted claims (over one year) is very high, even sometimes exceeding 1.5 predicted claims.
To better understand this, it would be interesting to carry a more in-depth study via local methods such as LIME and SHAP for example.\\

\textbf{Relevance according to the target audience} \\
Just like global methods, the analysis of the interactions is an essential phase for the actuary: firstly to understand the fitted model and ensure that is consistent with business knowledge and expertise but also to detect new relationships, like high order interaction.
On the side of the policyholder, this phase sounds less important because it will not be enough to understand why the model made such prediction. To have a simple explanation of the same type as the decomposition made for the GLM, local methods like LIME and SHAP seem necessary. 
However, interaction methods can be useful to explain to the policyholder that the effect of each feature is not linear and combined effect can exist. We could mention for example the well-know combined effect of the vehicle's power and the driver's age, which comes from an increased risk for a young driver with a powerful car.

\paragraph{\textbf{Local analysis via LIME and SHAP.}}

Global analysis tools (as the one introduced previously) provide useful insights on the average behaviour of the algorithm: the importance of each feature can be understood with the PFI, main effect of each feature on the prediction can be illustrated with PDP and ALE, main interactions detected with $H-$ statistics and magnitude of those interaction quantified with ICE or grouped PDP. However, global analysis might not be enough when it comes to detail an underwriting decision to an applicant that relies on a model.
Two interpretation \textbf{post-hoc} tools like LIME and SHAP (introduced previously) can then be useful.
Let us illustrate their usage through the example of a potential policyholder described above who has a frequency of claims predicted by the XGBoost of 94 \%.

\begin{figure}[!h]
    \centering
    \includegraphics[scale = 0.80]{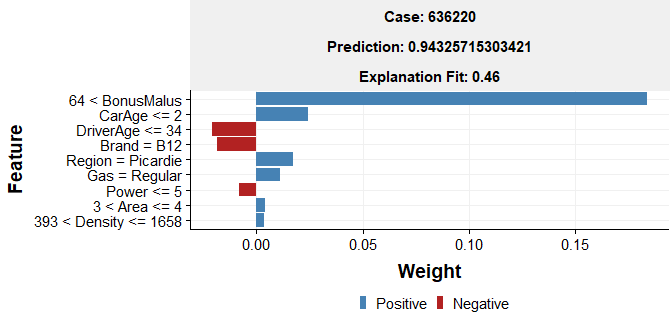}
    \caption{LIME features weight for a specific policyholder, based on the XGBoost model}
    \label{fig:LIME_xgb}
\end{figure}

Figure \ref{fig:LIME_xgb} shows the results provided by LIME method\footnote{with the following LIME parameters : $10 000$ permutations, all features (9) are used to fit the local surrogate model and the Gower distance function is used to assign a weight for each instance according to their distances to the point of interest.} for this policyholder, modeled by the XGBoost algorithm.

\begin{figure}[!h]
    \centering
    \includegraphics[scale = 0.70]{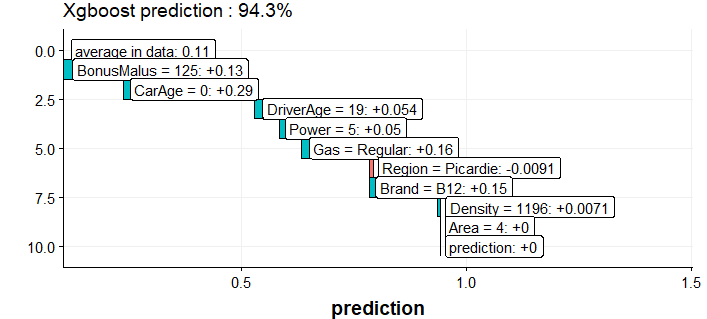}
    \caption{SHAP features contribution for a specific policyholder, based on the XGBoost model}
    \label{fig:SHAP_xgb}
\end{figure}

In complement to LIME analysis, Figure \ref{fig:SHAP_xgb} shows the results provided by SHAP method\footnote{with the following SHAP parameters : $500$ SHAP, $12$ permutations of random visit sequences, $4$ as a maximum number of rows in data to consider in the reference data.} for the same specific policyholder, modeled by the XGBoost model. 

Note that other parameters have been tested for both methods, and the results were not significantly impacted.

For these two methods, most of the features contributions are positive, which is consistent with the high prediction made by the model.
With LIME, the value of the feature Bonus Malus is the main reason of the XGBoost high prediction, which is also the case, but not to the same degree, with the CarAge and the Region.

With SHAP, we can see that the \textit{CarAge} is the most contributive feature, with \textit{Gas}, \textit{Band} and \textit{Bonus Malus}.

Even if the results provided by LIME and SHAP are substantially different, which is consistent because the underlying theories are different, these methods agree on the fact that Bonus Malus and Car Age have both a significant impact on the prediction, unlike the Driver Age which has little local contribution.

Global analysis showed that, on average, Driver Age has at a same time a huge importance in the model (in the top 3 most important features, see Figure \ref{fig:imp_var_dalex}), and a high positive contribution on the prediction when its value is 19 (see Figure \ref{fig:pdp_driver_car_age_xgb_xgb_cat}). LIME and SHAP analysis showed that in this local context, its role on the prediction was not so important and "contradicts" global analysis.
\\

\textbf{Relevance according to the target audience} \\
It seems clear that for a policyholder who wants to understand his tariff based on a complex model, the LIME and SHAP methods are the most appropriate interpretation methods.
Just like GLM model with coefficients (see Figure \ref{fig:prix_glm}), it allows to break down the prediction with a contribution of each feature (or for selected features with LIME).

On the side of the actuary, these methods applied on a single policyholder are not necessarily a knowledge he needs, but it can help him to understand if he can trust the model or not, or to discover new relationships.
Moreover, these methods applied on specific policyholders, for example with high or low predictions, or on a group of similar individuals can provide useful information. This can, for example, help to understand why a model will penalize or benefit certain specific profiles more, which can have negative consequences for the insurer.

\section*{Conclusion}
Predictive algorithms are increasingly popular within the insurance industry to improve the various of services it offers. Sometimes less restrictive in terms of assumptions to be verified than more traditional statistical algorithms, machine learning models (\textit{deep learning}, set methods, etc.) make it possible to improve the understanding of phenomena and to anticipate its occurrence with better precision. This gain is often made to the detriment of the complexity of the algorithms but also of the data on which they are based. The methods of interpretation of the algorithms make it possible to extend the analysis tools of these models in order to ensure their control and above all, their business relevance. Being able to understand and explain are essential elements for disciplines based on data manipulation. Thus, this article highlights certain methods of assistance in the interpretation of algorithms. It is important to stress that research continues to advance on this subject both from a technical and ethical point of view. Moreover, the tools introduced in this article are just some techniques among others with their advantages and limitations. The usage of only one \textit{post-hoc} metholodogy does not necessarily make an algorithm transparent. As most actuaries are used to, understanding or explaining a model lead to being able to control the mechanism of the algorithm, its dependency to the data (from which algorithms can help to emphasize potential bias for instance) and the results it can provide. Finally, transparency post-hoc tools are just additional algorithms and layers that bring more easy-to-read elements to complex, but more accurate, algorithms that need to be carefully used with their limitations in mind. Usage of several model analysis methods (as those introduced) help in lighting through different angles and reduce black spots on model decisions.

\begin{acknowledgements}
%If you'd like to thank anyone, place your comments here
%and remove the percent signs.
We are very grateful to the anonymous referees for valuable comments,
\end{acknowledgements}

% Authors must disclose all relationships or interests that 
% could have direct or potential influence or impart bias on 
% the work: 
%
\section*{Conflict of interest}
 The authors declare that they have no conflict of interest.

% BibTeX users please use one of
%\bibliographystyle{spbasic}      % basic style, author-year citations
%\bibliographystyle{spmpsci}      % mathematics and physical sciences
%\bibliographystyle{spphys}       % APS-like style for physics

%requested for Arxiv
%\bibliographystyle{abbrvnat}

\appendix
%TO translate
% \input{Librairies}
\section{Appendix: some useful open source libraries}
\label{libsOpen}

We list here some useful libraries that the reader can access in order to manipulate the presented methods.

\paragraph{R Packages :}
\begin{itemize}
    \item LIME : https://cran.r-project.org/web/packages/lime/index.html;\\ https://cran.r-project.org/web/packages/iml/index.html
    \item SHAP : https://github.com/slundberg/shap 
    \item H-statistics : https://cran.r-project.org/web/packages/iml/index.html
    \item ALE, PDP, ICE : https://cran.r-project.org/web/packages/iml/index.html;\\ https://cran.r-project.org/web/packages/pdp/index.html;\\ https://www.rdocumentation.org/packages/ICEbox/versions/1.1.2
    \item The importance of variables :
    https://cran.r-project.org/web/packages/iml/index.html;\\
    https://cran.r-project.org/web/packages/DALEX/index.html
\end{itemize}

\paragraph{Python Packages :}
\begin{itemize}
    \item LIME : https://github.com/marcotcr/lime
    \item SHAP : https://github.com/slundberg/shap 
    \item H-statistique : https://pypi.org/project/sklearn-gbmi/
    \item ALE, PDP, ICE : https://github.com/SauceCat/PDPbox; \\ 
    https://scikit-learn.org/ 
\end{itemize}

 \section{Appendix: PDP Curve Construction Algorithm}

\label{algo_pdp}
The algorithm proposed in \cite{pdp} to estimate the values taken by the partial dependency function is the following :

\begin{itemize}
    \item Input : the learning base $(x_j^{(i)})_{1 \leq i \leq n, 1 \leq j \leq p}$, the model $\hat{f}$, a variable to be explained assumed here to be $x_1$ to simplify, i.e. $S=\{1\}$ and $C=\{2,...,p\}$. \\
    That is: $(x_j^{(i)})_{1 \leq i \leq n, 1 \leq j \leq p} = (x_1^{(i)},x_C^{(i)})_{1 \leq i \leq n}$.
    \item for $i=1,...,n$:
    \begin{itemize}
        \item Copy the learning base, replacing the value of the variable $x_1$ by the constant value $x_1^{(i)}$: $(x_1^{(i)},x_C^{(k)})_{1 \leq k \leq n}$.
        \item Calculation of the prediction vector by $\hat{f}$ of the previously defined data: $\hat{f}(x_1^{(i)},x_C^{(k)})$ for $k=1,...,n$.
        \item Calculation of $\hat{f}_{x_1}(x_1 ^{(i)})$ by the formula: $\hat{f}_{x_1}(x_1^{(i)}) \simeq \frac{1}{n} \sum\limits_{k=1}^{n}{\hat{f}(x_1^{(i)},x_C^{(k)})} $
    \end{itemize}
    \item Output: the dot plot $(x_1^{(i)},\hat{f}_{x_1}(x_1^{(i)}))$ for $i=1,...,n$, called partial dependency graph (PDP).
\end{itemize}

 \section{Appendix: Precision on the ICE method}
\label{ICE_methodo}
The ICE curve approach provides a graph with one line for each instance, which shows how the prediction changes when a characteristic changes. 
Instead of the average performed in the PDP calculation, the ICE calculation is performed for each instance. This results in an ICE graph, containing as many curves as there are observations. This method has been introduced by \cite{ice}. Contrary to the partial dependency graph which is a global approach, the ICE curves are local (c.f figure \ref{scope_interpretability}). 
The algorithm used to estimate the ICE is the following:

\begin{itemize}
    \item Input: the training data:
    $(x^{(i)}_j)_{1 \leq i \leq n,1 \leq j \leq p}$, the adjusted model $\hat{f}$, $S$ a subset of $\{1,...p\}$ and $C$ the complementary of $S$ in $\{1,...,p\}$. 
    \item For i=1,...,n:
    \begin{itemize}
        \item $\hat{f}^{(i)}_S=0_{n\times1}$
        \item $x_C=x_C^{(i)}$: we set the $C$ columns to the $i$-th observation.
        \item For l=1,...n:
        \begin{itemize}
            \item $x_S=x_S^{(l)}$
            \item $\hat{f}^{(i)}_{S,l}=\hat{f}(x_S,x_C)$
        \end{itemize}
    \end{itemize}
    \item Output: $\hat{f}^{(1)}_{S}=(\hat{f}^{(1)}_{S,1},...,\hat{f}^{(1)}_{S,n}),...,\hat{f}^{(n)}_{S}=(\hat{f}^{(n)}_{S,1},...,\hat{f}^{(n)}_{S,n})$
\end{itemize}

Let's take again the illustrative example of the PDP method page \pageref{fig:scatterPDP}. In this example, we have observed that PDP, by realizing a mean, does not capture all the dependency of a variable on the prediction. 
Let us now display the ICE graph associated with the PDP.
\begin{figure}[!h]
    \centering
    \includegraphics[scale = 0.4]{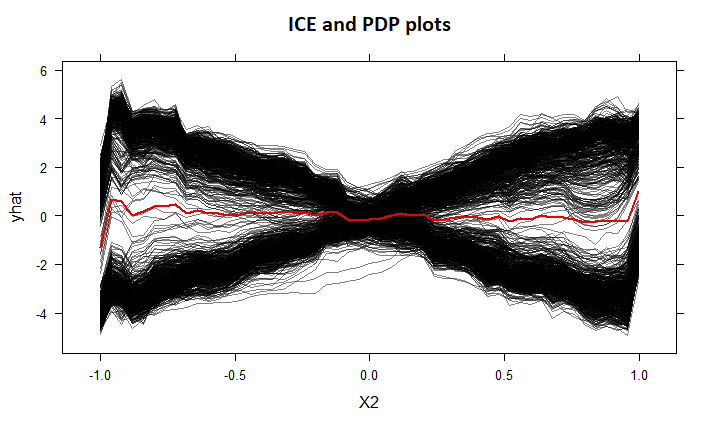}
    \caption{PDP graph (in red) and ICE curves (in black))}
    \label{fig:ice_pdp}
\end{figure}
This time, the ICE captures the effect of $X_2$ on the prediction for each instance. On average the ICE graph is close to 0, which corresponds to the PDP, but the full ICE graph is close to the scatter plot shown in figure \ref{fig:scatterPDP}.
There is an adaptation of the ICE method, called c-ICE, which allows to center the curves and thus to better see the effects of each variable on the prediction. Other variants based on partial derivatives such as d-ICE also exist.\\

The ICE curves are generally of the same interest as the PDP graph, notably the simplicity of implementation and interpretation. The other important advantage is that they do not mask the heterogeneous effects of the model under consideration. Thus, by using the PDP graph, which provides a summary of the impact of a variable on the model's prediction, and the ICE curves, which specify it, we obtain a good global explanation of the predictions.\\

However, like the PDP, the ICE curves are based on an assumption of independence between the different variables and do not take into account their actual distribution.
Another disadvantage is the fact that the curves can only be represented in 2 or 3 dimensions, because humans do not know how to represent higher dimensions. 
Moreover, the graph containing all the ICE curves is quickly overloaded when the number of individuals studied is large.

\section{Appendix : The ALE method }
\label{ALE_methodo}
To illustrate ALE, detailed in \cite{ale_pdp}, one could start by reminding the PDP formula.
PDP associated to variables $X_S$ lays on the following: \begin{equation}\hat{f}_{X_S,PDP}(x_S)=\mathds{E}_{X_C}[\hat{f}(x_S,X_C)]=\int\limits_{x_C}\hat{f}(x_S,x_C)\mathds{P}_{X_c}(x_C)dx_C\end{equation} for each point $x_S$ of the marginal distribution of $X_S$.
What we call $M$-plot, lays on the conditioned average of predictions:  
\begin{equation}\hat{f}_{X_S,M-plot}(x_S)=\mathds{E}_{X_C|X_S}[\hat{f}(X_S,X_C)|X_S=x_S]=\int\limits_{x_C}\hat{f}(x_S,x_C)\mathds{P}(x_C|x_S)dx_C\end{equation}

Finally, to produce ALE curve, we define interval bounds $z_{0,1}<x_S$ to compute the average of the difference of predictions. Formula then becomes:
\begin{equation}\hat{f}_{X_S,ALE}(x_S)=\int\limits_{z_{0,1}}^{x_S}{\mathds{E}_{X_C|X_S}[\hat{f}^{(S)}(X_S,X_C)|X_S=z_S]dz_S}=\int\limits_{z_{0,1}}^{x_S}\int\limits_{x_C}\hat{f}^{(S)}(x_S,x_C)\mathds{P}(x_C|z_S)dx_Cdz_S\end{equation}
where $\hat{f}^{(S)}(x_S,x_C)=\frac{\partial \hat{f}(x_S,x_C)}{\partial x_S}$ is gradient of $\hat{f}$ related to the first coordinate.

The ALE method differs from the PDP method, which directly averages predictions, in that it
calculates the prediction differences conditional on features S and integrates the derivative over features S to estimate the effect.
 Heuristically, the idea is that the derivation and integration cancel each other out and the derivative (or interval difference) isolates the effect of the feature of interest and blocks the effect of correlated features.

\paragraph{ALE estimation}

Let's zoom on the construction of the ALE for one unique numerical variable $x_j$ ($S=\{j\}$) with $j\in \mathds{N}$.
 We divide the set of values taken by the variable $x_j$ into several intervals: $[z_{0,j},z_{1,j}]$, $[z_{1,j},z_{2,j}]$,...,$[z_{k_j(x)-1,j},z_{k_j(x),j}]$, with $k_j(x)$ the number of intervals  $z_{0,j}<z_{1,j}<...<z_{k_j(x),j}$.
For $k \in \{1,...,k_j(x)\}$,  let $N_j(k)$ the set of individuals of the learning base for which the variable $x_j$ is in the interval number k: $[z_{k-1,j},z_{k,j}]$, and $n_j(k)$ the number of individuals in $N_j(k)$.
Then the ALE at the point $x\in \mathds{X}$ associated with the variable $j$ is estimated by the formula:
\begin{equation}
    \hat{\tilde{f}}_{j,ALE}(x)=\sum_{k=1}^{k_j(x)}\frac{1}{n_j(k)}\sum_{i:x_{j}^{(i)}\in{}N_j(k)}\left[f(z_{k,j},x^{(i)}_{\setminus{}j})-f(z_{k-1,j},x^{(i)}_{\setminus{}j})\right]
    \label{eq:ALE_1}
\end{equation}
The term "accumulated local effects" can be clearly explained on this formula: on each interval, we measure the "local" prediction difference, then we sum over all the intervals, in order to have the "accumulated" effect. In reality, the true ALE definition centers the previous term to have an average effect at 0, so the following formula follows:
\begin{align}
\hat{f}_{j,ALE}(x)&=\hat{\tilde{f}}_{j,ALE}(x)- \frac{1}{n}\sum\limits_{l=1}^{n}\hat{\tilde{f}}_{l,ALE}(x)\\ 
&=\sum_{k=1}^{k_j(x)}\frac{1}{n_j(k)}\sum_{i:x_{j}^{(i)}\in{}N_j(k)}\left[f(z_{k,j},x^{(i)}_{\setminus{}j})-f(z_{k-1,j},x^{(i)}_{\setminus{}j})\right] - \frac{1}{n}\sum\limits_{l=1}^{n}\hat{\tilde{f}}_{l,ALE}(x)
\label{eq:ALE_2}
\end{align}
Centering the formula allows the ALE to be interpreted as the effect of a variable on the prediction compared to the average prediction based on learning data.
For example, if we get an ALE of $-2$ at a certain point $x$ when $x_j=3$, this means that when the j-th variable is 3, the prediction value is 2 less than the average prediction. 

The intervals in the formula (\ref{eq:ALE_1}) are usually chosen as being different quantiles of the distribution of the variable $x_j$ under consideration. This allows to have as many individuals in each interval, but has the disadvantage of having intervals of very different sizes, especially in the case where the tail of the distribution is heavy.

 \section{Appendix: Alternatives to LIME}

\label{alternatives_LIME}
\subsection{Reminder on LIME}
The LIME procedure to explain the prediction made by a black box model, denoted $b$, for the $x$ instance is based on the steps detailed in the following algorithm:

\begin{itemize}
    \item The learning base $X_{s_x}$ to calibrate the substitution model $s_x$ is constructed by simulating samples of a normal distribution for each explanatory variable $x_i$, with the same mean and standard deviation as the variables used to fit $b$.
   We note for any $j \in \{1,... ,p\}, \mu_j=\frac{1}{n}\sum_{i=1}^{n}x_{ij}$ (empirical mean of the variable $x_j$ in the initial learning base) and $\sigma_j^2=\frac{1}{n- 1}\sum_{i=1}^{n}{(x_{ij}-\mu_j)^2}$ (empirical variance of the variable $x_j$ in the initial learning base). We also note $n_{sim}$ the number of individuals used to adjust $s_x$. Then for any $j \in \{1,...,p\}$, we simulate $n_{sim}$ independent samples of law $N(\mu_j,\sigma_j^2)$. This forms our base $X_{s_x}$. In the case of categorical variables, LIME works with the prediction probabilities returned by $b$.
   \item The substituted model $s_x$ is fitted using a linear model with ridge regularization, i.e. a penalty parameter is added to control the variance of the model. 
   Each instance $\tilde{x}$ of $X_{s_x}$ is given a weight calculated with respect to its distance with $x$, using a kernel function like the kernel $RBF$ defined by $RBF(x,\tilde{x})=exp(-\frac{||x-\tilde{x}|^2}{2 \sigma^2})$, where $\sigma > 0$ is a parameter to be defined.
   This weighting is intended to favour points close to the $x$ point that we are trying to explain, in order to have a local interpretation.
   \item Explanations of the black box $b$ are generated by extracting the coefficients of the linear regression implemented with the $s_x$ model.
\end{itemize}

\subsection{LIVE}
The article \cite{BreakDownLive} proposes two alternatives to the LIME and SHAP methods, called Live and BreakDown.
We focus in this part on the Live method (\emph{Local Interpretable Visual Explanations}).
The main objective of this method is the same as LIME, namely to explain a particular prediction of a black-box model using a local substitution model. The major difference with LIME lies in the choice of observations to fit the substitution model. The algorithm for generating the neighborhood is as follows:
\begin{itemize}
    \item Input: $n_{sim}$, the number of observations to generate and p, the number of explanatory variables used in the $b$ black box model. Let $x^* = (x^*_j)_{1\leq j\leq p}$ be the observation of interest that we want to explain. 
    \item Initialization: duplicate the given observations $n_{sim}$ times. Let us note $X'$ the matrix containing the learning data of the substitution model : 
$X' = \underbrace{
\begin{pmatrix}
    x^*_1      & \cdots & x^*_1 \\ 
    \vdots &  & \vdots \\ 
    x^*_p      & \cdots & x^*_p 
\end{pmatrix}}_{n_{sim} \text{times}}$.

    \item For $i \in \{1,...,n_{sim}\}$: Choose $k \in \{1,...,p\}$ randomly according to the uniform law: $k\sim \mathscr{U}([\![1,n_{sim}]\!])$. Replace the $k$ i-th variable with a draw of the empirical distribution (in the learning base) of this variable: $X'[k,i] \sim \mathscr{L}(X_k)$.

\end{itemize}
The notation $\mathscr{L}(X_k)$ corresponds to the "estimated" law of the variable $X_k$ which is approached either by a normal law (parameters found by the method of moments for example), or by random drawing of this variable on the basis of learning, which amounts to performing permutations on the initial basis. Both methods have the disadvantage of not taking into account the dependency between the explanatory variables. Alternatives exist but are not discussed here.
All the observations (of $X'$) thus generated by the above algorithm are assumed to be equidistant to the original observation that we are trying to explain: no weighting is applied when fitting the local substitution model. This frees us from the constraint of the choice of kernel of the LIME method.

\section{Appendix: Other graphs useful for analysis}
\label{annexe:graphes}

\begin{figure}[!h]
    \centering
    \includegraphics[scale = 0.17]{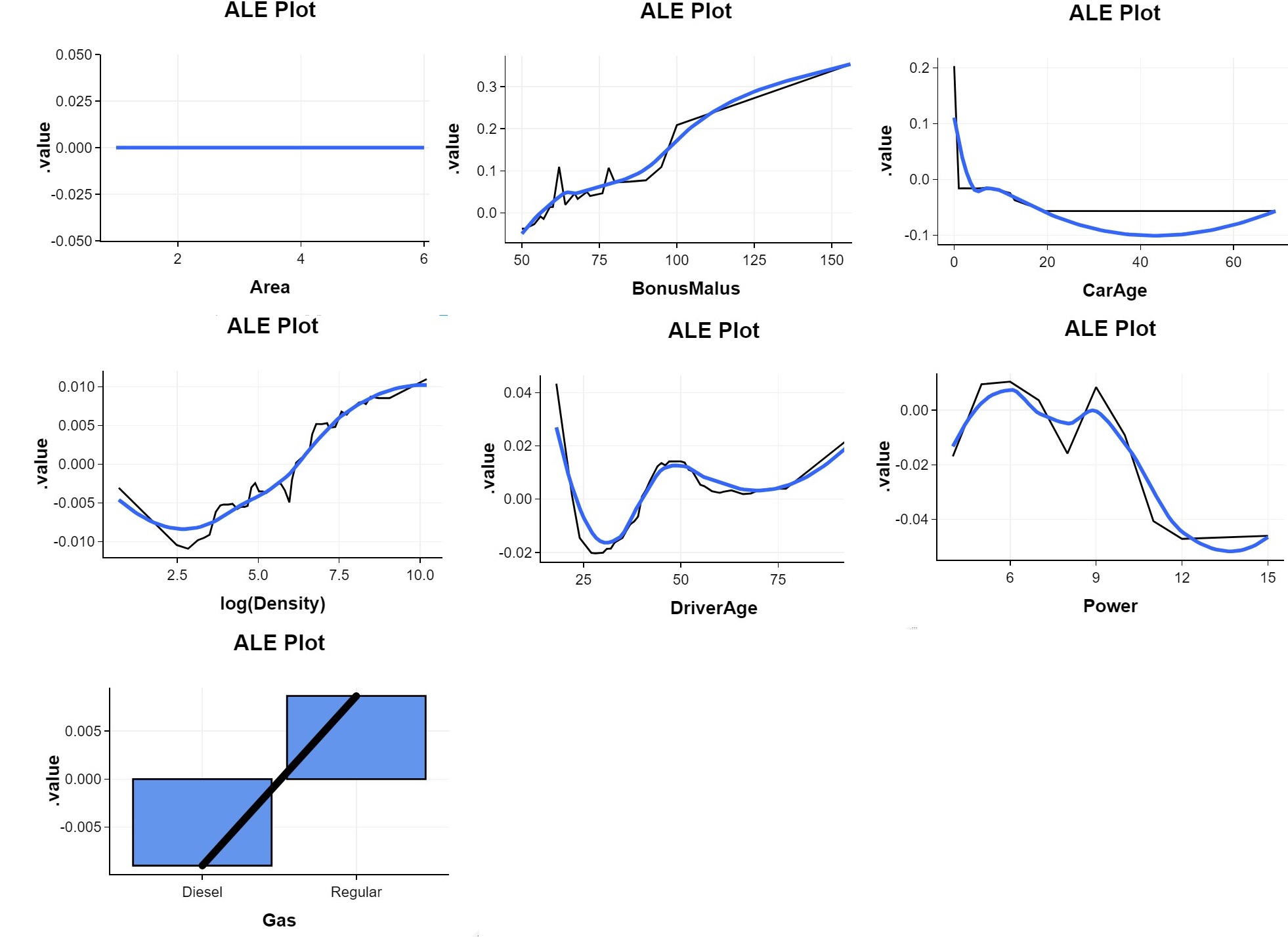}
    \caption{Accumulated Local Effect ($ALE$) associated with the different variables of the XGBoost $C$ model (in black the real ALE curve and in blue the smooth curve).}
    \label{fig:pdp_all_xgb_freq_num}
\end{figure}


\begin{thebibliography}{56}
\providecommand{\natexlab}[1]{#1}
\providecommand{\url}[1]{\texttt{#1}}
\expandafter\ifx\csname urlstyle\endcsname\relax
  \providecommand{\doi}[1]{doi: #1}\else
  \providecommand{\doi}{doi: \begingroup \urlstyle{rm}\Url}\fi

\bibitem[Aas et~al.(2019)Aas, Jullum, and Loland]{SHAP_dependant}
K.~Aas, M.~Jullum, and A.~Loland.
\newblock Explaining individual predicitions when features are dependent : More
  accurate approximations to Shapley values.
\newblock \emph{arXiv:1903.10464}, 2019.

\bibitem[Adadi and Berrada(2018)]{Peeking_Black_Box_XAI}
A.~Adadi and M.~Berrada.
\newblock Peeking inside the black-box: A survey on explainable artificial
  intelligence (xai).
\newblock \emph{IEEE Access}, 6, 2018.


\bibitem[Alvarez-Melis et~al. (2018)]{Alvarez}
D.~Alvarez-Melis and T.~Jaakkola.
\newblock On the robustness of interpretability methods.
\newblock \emph{In~: Proceedings of the 2018 ICML Workshop in Human Interpretability in Machine Learning}.
\newblock \emph{arXiv:1806.08049}, 2018.

\bibitem[Alvarez-Melis et~al. (2018)]{Alvarez1}
D.~Alvarez-Melis and T.~Jaakkola.
\newblock Towards robust interpretability with self-explaining neural networks.
\newblock \emph{In~: 32nd conference on Neural Information Processing Systems (NeurIPS 2018}.
\newblock \emph{arXiv:1806.07538}, 2018.


\bibitem[Andrews et~al.(1995)]{first_inter}
R.~Andrews, J.~Diederich and A.~B.~Tickle
\newblock Survey and critique of techniques for extracting rules from trained artificial neural networks.
\newblock \emph{Knowledge-Based Systems}, 8, 1995.

\bibitem[Apley(2014)]{ale}
D.W Apley.
\newblock Visualizing the effects of predictor variables in black box
  supervised learning models.
\newblock \emph{arXiv:1612.08468}, 2014.

\bibitem[Apley(2016)]{ale_pdp}
D.W Apley.
\newblock \emph{Visualizing the Effects of Predictor Variables in Black Box
  Supervised Learning Models}.
\newblock 2016.

\bibitem[D.~P.~Kingma(2017)]{adam_algo}
J.~L.~Ba and D.~P.~Kingma.
\newblock \emph{ADAM : A Method For Stochastic Optimization}.
\newblock 2017.

\bibitem[Bartlett et~al.(2017)Bartlett, Foster, and
  Telgarsky]{margin_bound_neural_network}
P.~Bartlett, D.~Foster, and M.~Telgarsky.
\newblock \emph{Spectrally-normalized margin bounds for neural networks}.
\newblock 2017.

\bibitem[Biecek and Staniak(2018)]{BreakDownLive}
P.~Biecek and M.~Staniak.
\newblock Explanations of model predictions with live and breakdown packages.
\newblock \emph{arXiv:1804.01955}, 2018.

\bibitem[Breiman et~al. (1996)Breiman, L., and S.]{Breiman1996}
L.~Breiman and N. Shang.
\newblock Born again trees.
\newblock  1996.

\bibitem[Breiman(2001)]{Breiman2001}
L.~Breiman.
\newblock Random forests.
\newblock \emph{Machine Learning}, 2001.

\bibitem[Buchner et~al.(2017)Buchner, J., and S.]{trees_interac}
F.~Buchner, Wasem J., and Schillo S.
\newblock Regression trees identify relevant interactions: can this improve the
  predictive performance of risk adjustment?
\newblock \emph{Health economics}, 2017.

\bibitem[Chen and Guestrin(2016)]{xgboost}
T.~Chen and C.~Guestrin.
\newblock Xgboost: {A} scalable tree boosting system.
\newblock \emph{CoRR}, abs/1603.02754, 2016.

\bibitem[Cook and Weisberg(1982)]{influence_function_cook_math}
R.~Cook and S.~Weisberg.
\newblock \emph{Residuals and influence in regression}.
\newblock New York: Chapman and Hall, 1982.

\bibitem[Craven et~al.(1995)]{trepan}
M.W.~Craven and J.~Shavlik.
\newblock Extracting tree-structured representations of trained networks.
\newblock \emph{NIPS}, 1995.

%\bibitem[Dalayan()]{app_stat_ensae}
%A.~S. Dalayan.
%\newblock \emph{Cours d'Apprentissage et Data Mining (Ensae ParisTech)}.

\bibitem[Delcaillau(2019)]{Delcaillau}
D.~Delcaillau.
\newblock Contr\^ole et transparence des mod\`eles complexes en actuariat.
\newblock \emph{M\'emoire de l'Institut des Actuaires}, 2019.



\bibitem[Doshi-Velez et~al(2017)]{Doshi}
F.~Doshi-Velez, and B.~Kim.
\newblock Towards a rigorous science of interpretable machine learning.
\newblock \emph{arXiv:1702.08608}, 2017.

\bibitem[Fisher(2018)]{variable_importance}
A.~Fisher.
\newblock \emph{All Models are Wrong but Many are Useful: Variable Importance
  for Black-Box, Proprietary, or Misspecified Prediction Models, using Model
  Class Reliance}.
\newblock 2018.

\bibitem[Freund and Schapire(1997)]{freund1997decision}
Y.~Freund and R.~Schapire.
\newblock A decision-theoretic generalization of on-line learning and an
  application to boosting.
\newblock \emph{Journal of computer and system sciences}, 55\penalty0
  (1):\penalty0 119--139, 1997.

\bibitem[Freund et~al.(1996)Freund, Schapire, et~al.]{freund1996experiments}
Y.~Freund, R.~Schapire, et~al.
\newblock Experiments with a new boosting algorithm.
\newblock In \emph{icml}, volume~96, pages 148--156, 1996.

\bibitem[Friedman et~al.(2008)Friedman, Popescu, et~al.]{H_stat}
J.~Friedman, B.~Popescu, et~al.
\newblock Predictive learning via rule ensembles.
\newblock \emph{The Annals of Applied Statistics}, 2\penalty0 (3):\penalty0
  916--954, 2008.

\bibitem[Friedman(2001)]{pdp}
J.H. Friedman.
\newblock Greedy function approximation : A gradient boosting machine.
\newblock \emph{The Annals of Statistics}, 2001.

\bibitem[Friedman(1940)]{friedman1940comparison}
M.~Friedman.
\newblock A comparison of alternative tests of significance for the problem of
  m rankings.
\newblock \emph{The Annals of Mathematical Statistics}, 11\penalty0
  (1):\penalty0 86--92, 1940.

\bibitem[Goldstein et~al.(2015)Goldstein, Kapelner, Bleich, and Pitkin]{ice}
A.~Goldstein, A.~Kapelner, J.~Bleich, and E.~Pitkin.
\newblock Peeking inside the black box: Visualizing statistical learning with
  plots of individual conditional expectation.
\newblock \emph{Journal of Computational and Graphical Statistics}, 24\penalty0
  (1):\penalty0 44--65, 2015.

\bibitem[Greenwell(2018)]{pdp_feature_imp}
B.~M. Greenwell.
\newblock A simple and effective model-based variable importance measure.
\newblock \emph{arXiv:1805.04755}, 2018.

\bibitem[I.~J.~Goodfellow(2015)]{adversial_examples}
C.~Szegedy I.~J.~Goodfellow, J.~Shlens.
\newblock \emph{Explaining and Harnessing Adversial Examples}.
\newblock 2015.

\bibitem[Kim(2016)]{prototypes_critiques}
B.~Kim.
\newblock \emph{Examples are not Enough, Learn to Criticize! Criticism for
  Interpretability}.
\newblock 2016.

\bibitem[Kim et~al.(2016)Kim, Khanna, and Koyejo]{NIPS2016_6300}
B.~Kim, R.~Khanna, and O.~Koyejo.
\newblock Examples are not enough, learn to criticize! criticism for
  interpretability.
\newblock In D.~D. Lee, M.~Sugiyama, U.~V. Luxburg, I.~Guyon, and R.~Garnett,
  editors, \emph{Advances in Neural Information Processing Systems 29}, pages
  2280--2288. Curran Associates, Inc., 2016.

\bibitem[Koh(2017)]{influence_function}
P.~W. Koh.
\newblock \emph{Understanding Black-box Predictions via Influence Functions}.
\newblock 2017.

\bibitem[Lange(2008)]{shapley_value}
F.~Lange.
\newblock \emph{Exploration de la valeur de Shapley et des indices
  d'interaction pour les jeux d\'efinis sur des ensembles ordonn\'es}.
\newblock 2008.

\bibitem[Laugel(2017)]{inverse_classification}
T.~Laugel.
\newblock \emph{Inverse Classification for Comparison-based Interpretability in
  Machine Learning}.
\newblock 2017.

\bibitem[Leroy and Planchet(2009)]{segmentationPlanchet}
G.~Leroy and F.~Planchet.
\newblock Quel niveau de segmentation pertinent ?
\newblock \emph{La Tribune de l'Assurance}, 2009.

\bibitem[Lipton(2018)]{mythos_interpretability}
Z.~Lipton.
\newblock The mythos of model interpretability.
\newblock \emph{Queue}, 16\penalty0 (3):\penalty0 31--57, 2018.

\bibitem[Lundberg and Lee(2017)]{shap}
S.~Lundberg and Su-In Lee.
\newblock A unified approach to interpreting model predictions.
\newblock pages 4765--4774, 2017.

\bibitem[Ly(2019)]{ly:tel-02413664}
A.~Ly.
\newblock Machine learning algorithms in insurance : solvency, textmining,anonymization and transparency.
\newblock \emph{HAL:tel-02413664}, \penalty0 (2019PESC2030), 2019.

\bibitem[Meinshausen(2010)]{meinshausen}
N.~Meinshausen.
\newblock Node harvest.
\newblock \emph{The Annals of Applied Statistics}, 4(4) :\penalty0 2049--2072, 2010.



\bibitem[Miller(2019)]{miller}
T.~Miller.
\newblock Explanation in artificial intelligence: Insights from the social
  sciences.
\newblock \emph{Artificial Intelligence}, 267:\penalty0 1--38, 2019.

\bibitem[Mohseni et al(2020)]{Mohseni}
S.~Mohseni, N.~Zaei, and E.D. Ragan
\newblock A multidisciplinary survey and framework for design and evaluation of explainable AI systems.
\newblock \emph{ACM Trans. Interact. Intell. Syst.}, 
1 (1), 2020.

\bibitem[Molnar(2019)]{molnar2019}
C.~Molnar.
\newblock \emph{Interpretable Machine Learning - A Guide for Making Black Box
  Models Explainable}.
\newblock 2019.

\bibitem[Murdoch and Singh(2019)]{define_interpretability_ML}
W.~J Murdoch and C.~Singh.
\newblock Interpretable machine learning: Definitions, methods, and
  applications.
\newblock \emph{arXiv:1901.04592}, 2019.

\bibitem[Noll et~al.(2020)Noll, Salzmann, and Wuthrich]{case_study}
A.~Noll, R.~Salzmann, and M.~Wuthrich.
\newblock Case study: French motor third-party liability claims.
\newblock \emph{Available at SSRN 3164764}, 2020.

\bibitem[Olden et~al.(2004)Olden, Joy, and Death]{olden2004accurate}
J.~Olden, M.~Joy, and R.~Death.
\newblock An accurate comparison of methods for quantifying variable importance
  in artificial neural networks using simulated data.
\newblock \emph{Ecological Modelling}, 178\penalty0 (3-4):\penalty0 389--397,
  2004.

\bibitem[Ribeiro et~al.(2016)Ribeiro, Singh, and Guestrin]{lime}
M.~Ribeiro, S.~Singh, and C.~Guestrin.
\newblock "why should I trust you?" explaining the predictions of any
  classifier.
\newblock pages 1135--1144, 2016.

\bibitem[Ribeiro et~al.(2018)Ribeiro, Singh, and Guestrin]{anchors}
M.~Ribeiro, S.~Singh, and C.~Guestrin.
\newblock \emph{Anchors: High-precision model-agnostic explanations}.
\newblock 2018.

\bibitem[Schapire(2013)]{schapire2013explaining}
R.~Schapire.
\newblock Explaining adaboost.
\newblock In \emph{Empirical inference}, pages 37--52. Springer, 2013.

\bibitem[Simonyan et~al.(2014)Simonyan, Vedaldi, and
  Zisserman]{deep_inside_convolutional_network}
K.~Simonyan, A.~Vedaldi, and A.~Zisserman.
\newblock \emph{Deep inside convolutional networks: Visualising image
  classification models and saliency maps}.
\newblock Iclr, 2014.

\bibitem[Strumbelj and Kononenko(2011)]{shap_fast}
E.~Strumbelj and I.~Kononenko.
\newblock A general method for visualizing and explaining black-box regression
  models.
\newblock \emph{Int. Conf. on Adaptive and Natural Computing Algorithms}, 2011.

\bibitem[Szegedy et~al.(2013)Szegedy, Zaremba, Sutskever, Bruna, Erhan,
  Goodfellow, and Fergus]{intriguing_neural_network}
C.~Szegedy, W.~Zaremba, I.~Sutskever, J.~Bruna, D.~Erhan, I.~Goodfellow, and
  R.~Fergus.
\newblock \emph{Intriguing properties of neural networks}.
\newblock 2013.

\bibitem[T.~Laugel(2018)]{limite_lime}
X.~Renard T.~Laugel.
\newblock \emph{Defining Locality for Surrogates in Post-hoc Interpretablity}.
\newblock 2018.

\bibitem[Tan et~al.(2019)Tan, Song, and Udell]{why_trust_LIME}
H.~F. Tan, K.~Song, and M.~Udell.
\newblock Why should you trust my interpretation? understanding uncertainty in
  lime predictions.
\newblock \emph{arXiv:1904.12991}, 2019.

\bibitem[Tibshirani(1990)]{tibshirani1990generalized}
R.~Tibshirani.
\newblock \emph{Generalized additive models}.
\newblock London: Chapman and Hall, 1990.

\bibitem[Vermet(2020)]{app_stat_vermet}
F.~Vermet.
\newblock \emph{Cours d'Apprentissage Statistique : une approche connexionniste
  (EURIA)}.
\newblock 2020.

\bibitem[Wachter et~al.(2017)]{explication_contrefactuelle}
S.~Wachter, B.~Mittelstadt, and C.~Russell.
\newblock \emph{Counterfactual explanations without opening the black box:
  Automated decisions and the GDPR}, volume~31.
\newblock HeinOnline, 2017.

\bibitem[Winter(2002)]{winter2002shapley}
E.~Winter.
\newblock The shapley value.
\newblock \emph{Handbook of game theory with economic applications},
  3:\penalty0 2025--2054, 2002.
  

\end{thebibliography}
\end{document}